\documentclass[sigconf, screen]{acmart}

\acmYear{}
\acmISBN{}
\acmDOI{}
\renewcommand\footnotetextcopyrightpermission[1]{}
\setcopyright{none}

\newcommand{\std}[1]{\textcolor{gray}{\scriptsize #1}}

\usepackage{balance} 
\usepackage{color}
\usepackage{xcolor}
\usepackage{graphicx}
\usepackage{subcaption}
\usepackage{array}

\definecolor{Dandelion}{RGB}{240, 225, 48}
\definecolor{Purple}{RGB}{160, 32, 240}
\definecolor{RoyalBlue}{RGB}{65, 105, 225}
\definecolor{ForestGreen}{RGB}{34, 139, 34}
\definecolor{Gray}{RGB}{190, 190, 190}
\definecolor{Maroon}{RGB}{176, 48, 96}
\definecolor{YellowGreen}{RGB}{154, 205, 50}

\definecolor{gray1}{RGB}{210,210,210}
\definecolor{gray2}{RGB}{50,50,50}

\definecolor{grasper}{HTML}{0000FF}
\definecolor{irrigator}{HTML}{FF0000}
\definecolor{bipolarforceps}{HTML}{FF00FF}
\definecolor{sealerdivider}{HTML}{00FFFF}
\definecolor{scissors}{HTML}{00FF00}
\definecolor{hook}{HTML}{FF8000}
\definecolor{suturinginstrument}{HTML}{FFFF00}
\definecolor{morcellator}{HTML}{7F2E03}
\definecolor{trocarsleeve}{HTML}{56BD89}
\definecolor{cannula}{HTML}{5957B9}
\definecolor{thread}{HTML}{74A200}
\definecolor{uterus}{HTML}{FFA07A}
\definecolor{tube}{HTML}{0580E4}
\definecolor{ovary}{HTML}{104B00}
\definecolor{shadecolor}{rgb}{0.97,0.97,0.97}

\usepackage{tikz}
\usepackage{multirow}
\usepackage{pifont}

\newcommand{\DP}[2]{%
  \begin{tikzpicture}
    \fill[color=#2]   (0.0 , 0.0) rectangle (#1*8.5ex , 2ex );
  \end{tikzpicture}%
}

\newcommand{\DPlegend}[2]{%
  \begin{tikzpicture}[baseline=(current bounding box.center)]
    \fill[color=#2] (0.0, 0.0) rectangle (2.2em, 2ex);  
  \end{tikzpicture}%
}

\AtBeginDocument{%
  }

\usepackage{calc}

\usepackage{stackengine}
\newcommand\clapp[3][0pt]{\stackengine{0pt}{#3}{\kern#1#2}{O}{c}{F}{F}{L}}

\begin{document}

\title{GynSurg: A Comprehensive Gynecology Laparoscopic Surgery Dataset}

\author{Sahar Nasirihaghighi}
\email{Sahar.Nasirihaghighi@aau.at}
\affiliation{%
  \institution{University of Klagenfurt}
  \city{Klagenfurt}
  \country{Austria}
}

\author{Negin Ghamsarian}
\email{negin.ghamsarian@unibe.ch}
\affiliation{%
  \institution{University of Bern}
  \city{Bern}
  \country{Switzerland}
}

\author{Leonie Peschek}
\email{Leonie-sophie.peschek@meduniwien.ac.at}
\affiliation{%
  \institution{Medical University of Vienna} 
  \city{Vienna}
  \country{Austria}
}

\author{Matteo Munari}
\email{matteo.munari@meduniwien.ac.at}
\affiliation{%
  \institution{Medical University of Vienna}
  \city{Vienna}
  \country{Austria}
}

\author{Heinrich Husslein}
\email{Heinrich.Husslein@meduniwien.ac.at}
\affiliation{%
  \institution{Medical University of Vienna}
  \city{Vienna}
  \country{Austria}
}

\author{Raphael Sznitman}
\email{raphael.sznitman@unibe.ch}
\affiliation{%
  \institution{University of Bern}
  \city{Bern}
  \country{Switzerland}
}

\author{Klaus Schoeffmann}
\email{Klaus.Schoeffmann@aau.at}
\affiliation{%
  \institution{University of Klagenfurt}
  \city{Klagenfurt}
  \country{Austria}
}

\renewcommand{\shortauthors}{Nasirihaghighi et al.}

\begin{abstract}
Recent advances in deep learning have transformed computer-assisted intervention and surgical video analysis, driving improvements not only in surgical training, intraoperative decision support, and patient outcomes, but also in postoperative documentation and surgical discovery. Central to these developments is the availability of large, high-quality annotated datasets. In gynecologic laparoscopy, surgical scene understanding and action recognition are fundamental for building intelligent systems that assist surgeons during operations and provide deeper analysis after surgery. However, existing datasets are often limited by small scale, narrow task focus, or insufficiently detailed annotations, limiting their utility for comprehensive, end-to-end workflow analysis. To address these limitations, we introduce GynSurg, the largest and most diverse multi-task dataset for gynecologic laparoscopic surgery to date. GynSurg provides rich annotations across multiple tasks, supporting applications in action recognition, semantic segmentation, surgical documentation, and discovery of novel procedural insights. We demonstrate the dataset’s quality and versatility by benchmarking state-of-the-art models under a standardized training protocol. To accelerate progress in the field, we publicly release the GynSurg dataset and its annotations (\url{https://ftp.itec.aau.at/datasets/GynSurge/}).

\end{abstract}


\maketitle

\section{Introduction}

Minimally invasive surgery (MIS)~\cite{tsui2013minimally}, commonly referred to as “keyhole” or endoscopic surgery, has revolutionized modern medicine by enabling clinicians to access internal anatomy through small incisions~\cite{schoeffmann2018video}. Compared to traditional open procedures, MIS significantly reduces patient trauma, accelerates postoperative recovery, and shortens hospital stays~\cite{kulkarni2020laparoscopic, aldahoul2021transfer}. Among MIS techniques, laparoscopy has become the gold standard for a wide range of abdominal interventions, including cholecystectomy and hysterectomy, due to its combination of high-resolution imaging and precise instrument manipulation.

In gynecologic laparoscopy, a minimally invasive approach for treating disorders of the female reproductive system, a rigid laparoscope is inserted into the abdominal cavity through a trocar port, transmitting high-definition video of the uterus, fallopian tubes, ovaries, and surrounding tissues to external monitors~\cite{lux2010novel, loukas2015smoke, leibetseder2019glenda}. These video recordings serve not only as a vital tool for intraoperative navigation but also as an invaluable resource for postoperative documentation, surgical education, and discovery, capturing detailed information on surgical maneuvers, tissue interactions, and surgeon expertise~\cite{munzer2013relevance, schoeffmann2015keyframe}. However, the sheer volume and complexity of laparoscopic video data far exceed the capacity for manual review and analysis~\cite{funke2019video, munzer2018content}.

As a result, Automated analysis of gynecologic laparoscopic videos has become an important tool for enhancing intraoperative assistance and postoperative evaluation, with the potential to improve patient outcomes, enhance surgical training, and streamline postoperative assessment~\cite{schoeffmann2015keyframe}. Prior research has addressed specific tasks such as surgical tool detection and localization~\cite{teevno2023semi, alshirbaji2021deep}, anatomical structure segmentation~\cite{madad2020surgai, den2023deep, tokuyasu2021development}, and the recognition of surgical actions~\cite{nasirihaghighi2023action, zhang2023laparoscopic} and events~\cite{nasirihaghighi2024event, kitaguchi2020real}. Nevertheless, the development of robust, generalizable models remains constrained by the limitations of existing public datasets, which are often small in scale, focused on a single task, or lacking comprehensive multimodal annotations. This gap underscores the urgent need for large-scale, richly annotated datasets to drive progress in intraoperative decision support, postoperative analysis, and surgical discovery.

\begin{table}[bt]
  \centering
  \caption{Comparison of publicly available gynecologic laparoscopic surgery datasets. GynSurg offers the most comprehensive coverage across multiple tasks and label types.}
  \label{tab:Comparison of publicly dataset}
  \small
  \resizebox{\columnwidth}{!}{%
    \begin{tabular}{lc*{3}{c}*{2}{c}}
      \toprule
        \textbf{Dataset} 
        & \textbf{Task} 
        & \textbf{Data Volume} 
        & \textbf{Annotations}\\
      \midrule

      SurgAI3.8K \cite{zadeh2023surgai3}   
        & \begin{tabular}[c]{@{}c@{}}Segmentation \\
        \end{tabular}
        & \begin{tabular}[c]{@{}c@{}}79 videos\\ (3,800 frames)\\
        \end{tabular}
        & \begin{tabular}[c]{@{}c@{}}2 Anatomical Structures \\
        \end{tabular}\\
        \midrule
      SurgAI \cite{madad2020surgai}   
        & \begin{tabular}[c]{@{}c@{}}Segmentation \end{tabular}
        & \begin{tabular}[c]{@{}c@{}}8 videos \end{tabular}
        & \begin{tabular}[c]{@{}c@{}}2 Anatomical Structures, \\ Surgical Instrument\end{tabular}\\
        \midrule
        
        \multirow{4}{*}{AutoLaparo \cite{wang2022autolaparo}}
          & Phase Recognition 
            & 21 videos 
            & 7 Surgical Phases \\
        \cmidrule[0.01pt](lr){2-4}
          & Motion Prediction 
            & 300 clips 
            & 7 Modes \\
        \cmidrule[0.01pt](lr){2-4}
          & \multirow{2}{*}{Segmentation}
            & \multirow{2}{*}{21 videos}
            & Anatomical Structure, \\
          & 
            & 
            & 4 Surgical Instruments \\
        \midrule
        
      GLENDA \cite{leibetseder2019glenda}  
        & \begin{tabular}[c]{@{}c@{}}Detection \\ and Localization\end{tabular}
        & \begin{tabular}[c]{@{}c@{}}25,682 frames \\ \end{tabular}
        & \begin{tabular}[c]{@{}c@{}}5 pathological Categories \\ \end{tabular}\\
        \midrule
      SurgicalActions160 \cite{schoeffmann2018video}    
        & Action Recognition 
        & 160 clips
        & 16 Actions\\
        \midrule
        
      GynLap4 \cite{leibetseder2018lapgyn4}
        & \begin{tabular}[c]{@{}c@{}}Classification \\ \end{tabular}
        & \begin{tabular}[c]{@{}c@{}}30,682 frames\\ 2,728 frames \\ 4,787 frames \\ 21,424 frames \\ \end{tabular}
        & \begin{tabular}[c]{@{}c@{}} 8 Surgical Actions,\\ 5 Anatomical Structures, \\ Actions on Anatomy, \\ Instrument Count\end{tabular}\\
        \midrule
        
      \multirow{3}{*}{GynSurg}
      & Action Recognition
        & \begin{tabular}[c]{@{}c@{}} 152 video cases \end{tabular}
        & \begin{tabular}[c]{@{}c@{}} 4 Surgical Actions,\\ 2 Side‐effects \end{tabular} \\
      \cmidrule[0.01pt](lr){2-4}
      & Segmentation
        & \begin{tabular}[c]{@{}c@{}}15 video cases\\(12,362 frames)\end{tabular}
        & \begin{tabular}[c]{@{}c@{}}21 Surgical Instruments,\\4 Anatomical Structures\end{tabular} \\
    \bottomrule
      
    \end{tabular}%
  }
\end{table}

\begin{table}[t!]
\centering
\caption{Overview of dataset composition across procedural actions and side-effect categories}
\label{tab:dataset_composition}
\resizebox{0.45\textwidth}{!}{%
  \begin{tabular}{%
      >{\raggedright\arraybackslash}m{2cm}   
      >{\raggedright\arraybackslash}m{2cm} 
      *{4}{>{\centering\arraybackslash}m{2cm}} 
    }
    \specialrule{.12em}{.05em}{.05em}
    \textbf{Category}
      & \textbf{Class Name}
      & \textbf{Video Case}
      & \textbf{Segment}
      & \textbf{Clip (3 sec)}
      & \textbf{Total duration (sec)} \\
    \midrule
    \multirow{5}{*}{Action}
      & NeedlePassing   & 48  & 510 & 1,206 & 7,036 \\
      & Coagulation      & 26  & 321 & 1,068 & 2,766 \\
      & Suction/Irrigation & 51 & 144 &  212 &  722 \\
      & Transection     &  8  &  69 &  354 &  848 \\
      & Rest            & 48  & 790 & 1,100 & 24,067 \\
    \midrule
    \multirow{5}{*}{Side effect}
      & Bleeding        & 47  & 114 &  977 & 2,240 \\
      & Non-bleeding    & 21  & 304 & 1,064 & 14,140  \\
    \cmidrule(lr){2-6}
      & Smoke           & 35  & 275 & 4,193 & 8,922 \\
      & Non-smoke       & 36  & 948 & 4,200 & 29,507 \\
    \specialrule{.12em}{.05em}{.05em}
  \end{tabular}%
}
\end{table}

\begin{figure}[b]
  \centering
  \begin{subfigure}[t]{0.32\columnwidth}
    \centering
    \includegraphics[width=\linewidth]{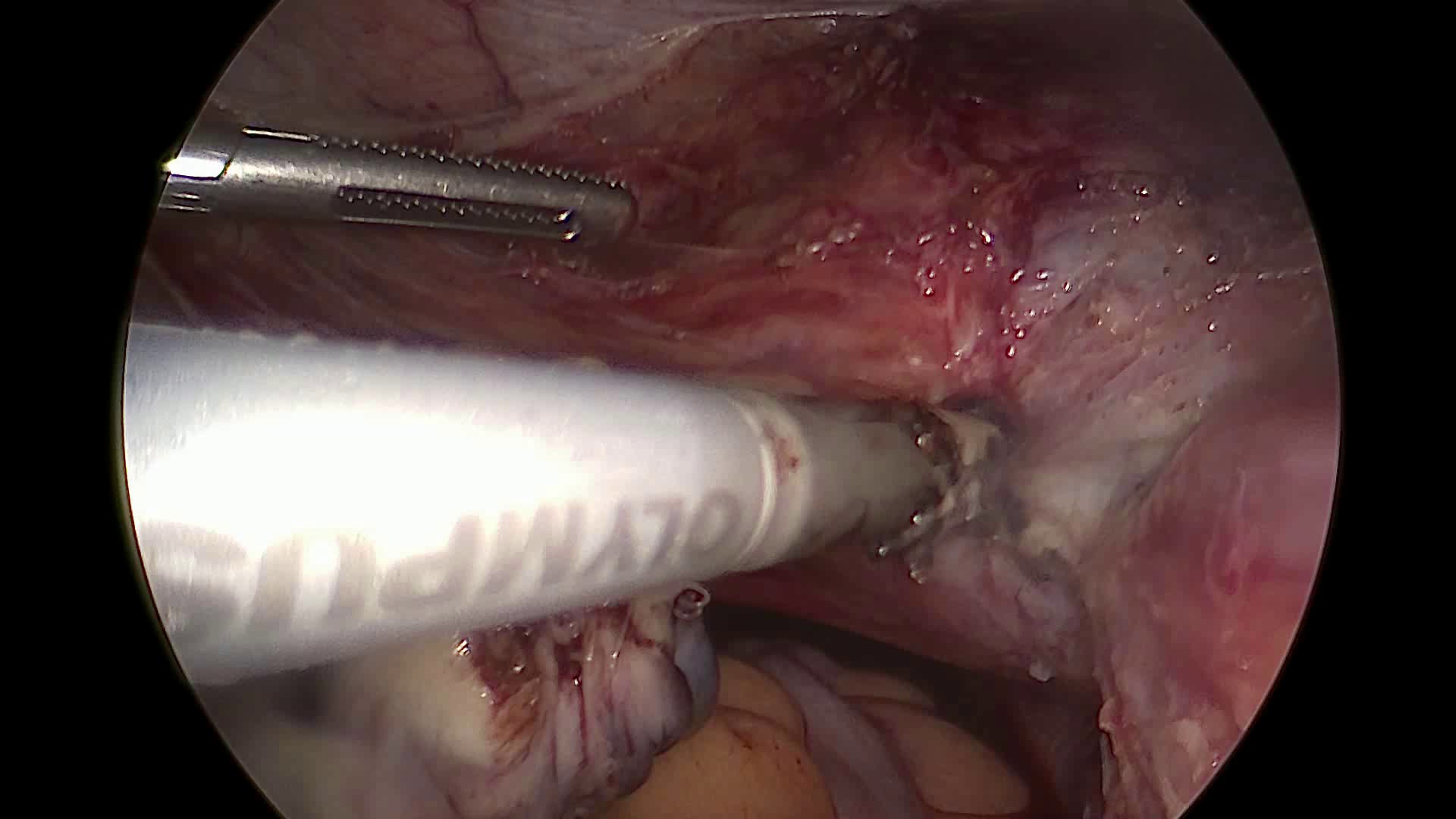}
    \caption{Coagulation}    
  \end{subfigure}
  \begin{subfigure}[t]{0.32\columnwidth}
    \centering
    \includegraphics[width=\linewidth]{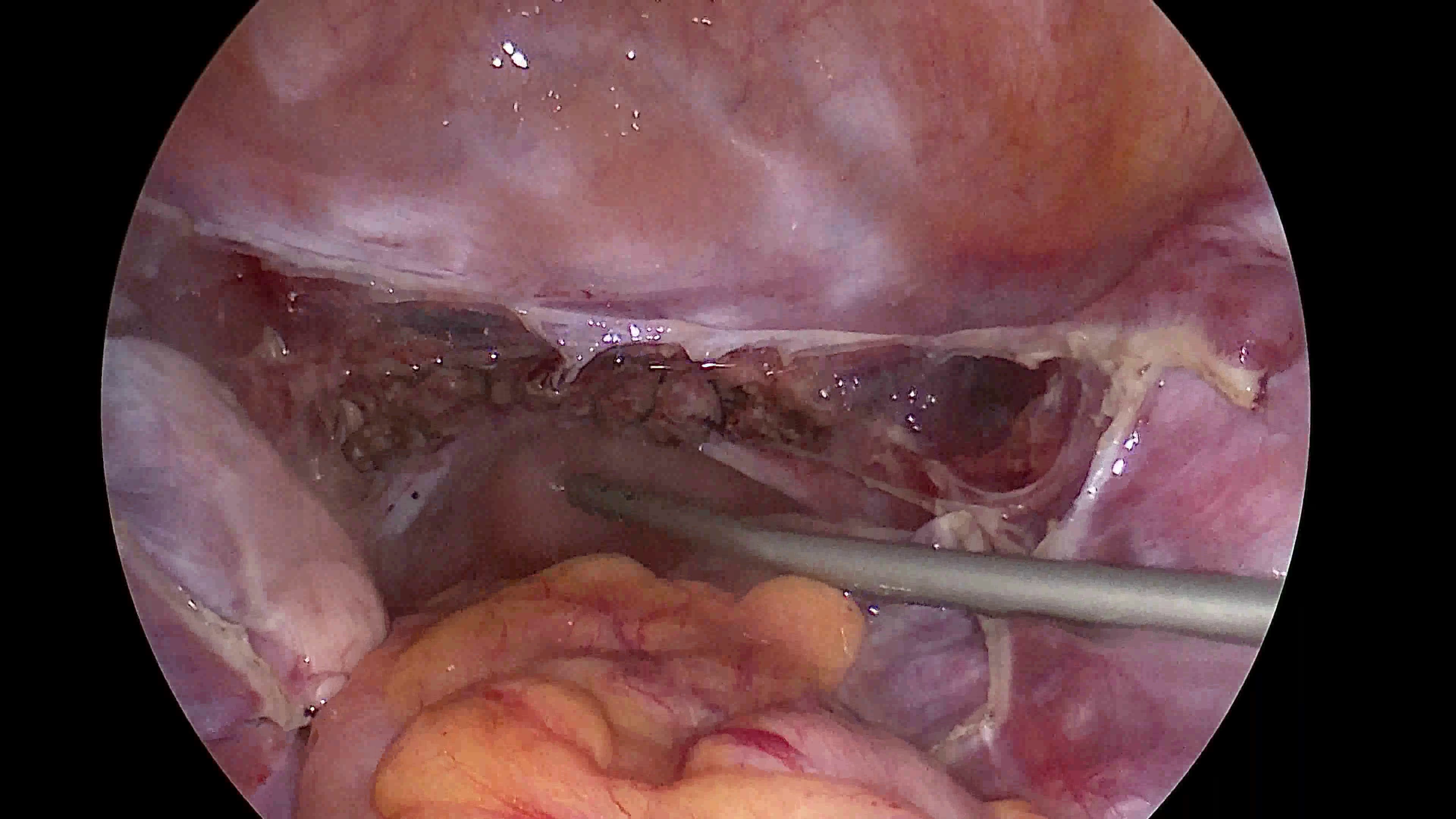}
    \caption{Suction/Irrigation}    
  \end{subfigure}
  \begin{subfigure}[t]{0.31\columnwidth}
    \centering
    \includegraphics[width=\linewidth]{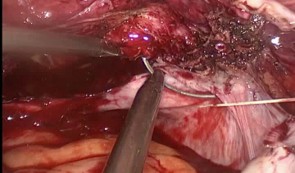}
    \caption{Needle Passing}    
  \end{subfigure}
  \\
  \begin{subfigure}[t]{0.32\columnwidth}
    \centering
    \includegraphics[width=\linewidth]{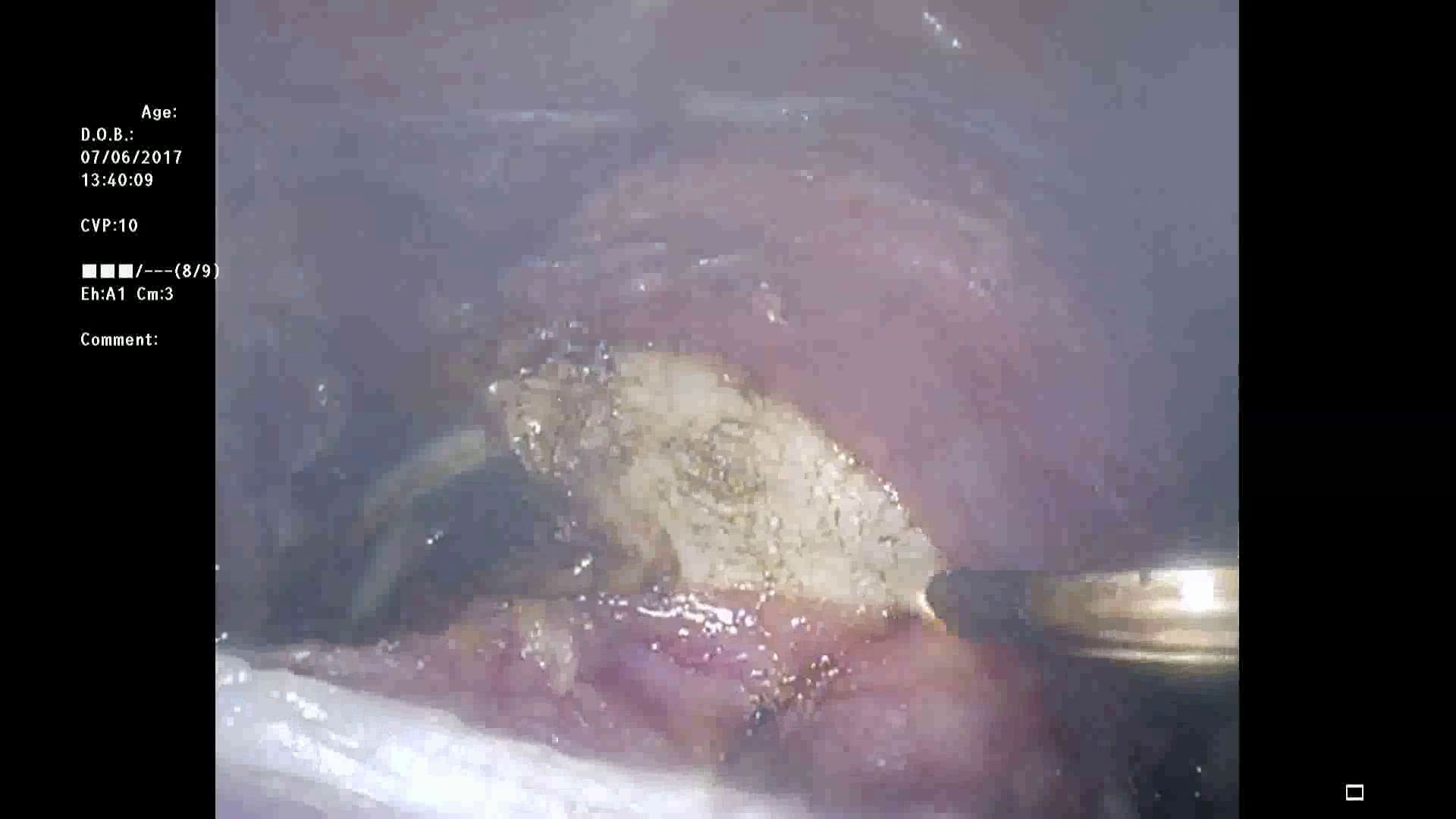}
    \caption{Transection}    
  \end{subfigure}
  \begin{subfigure}[t]{0.32\columnwidth}
    \centering
    \includegraphics[width=\linewidth]{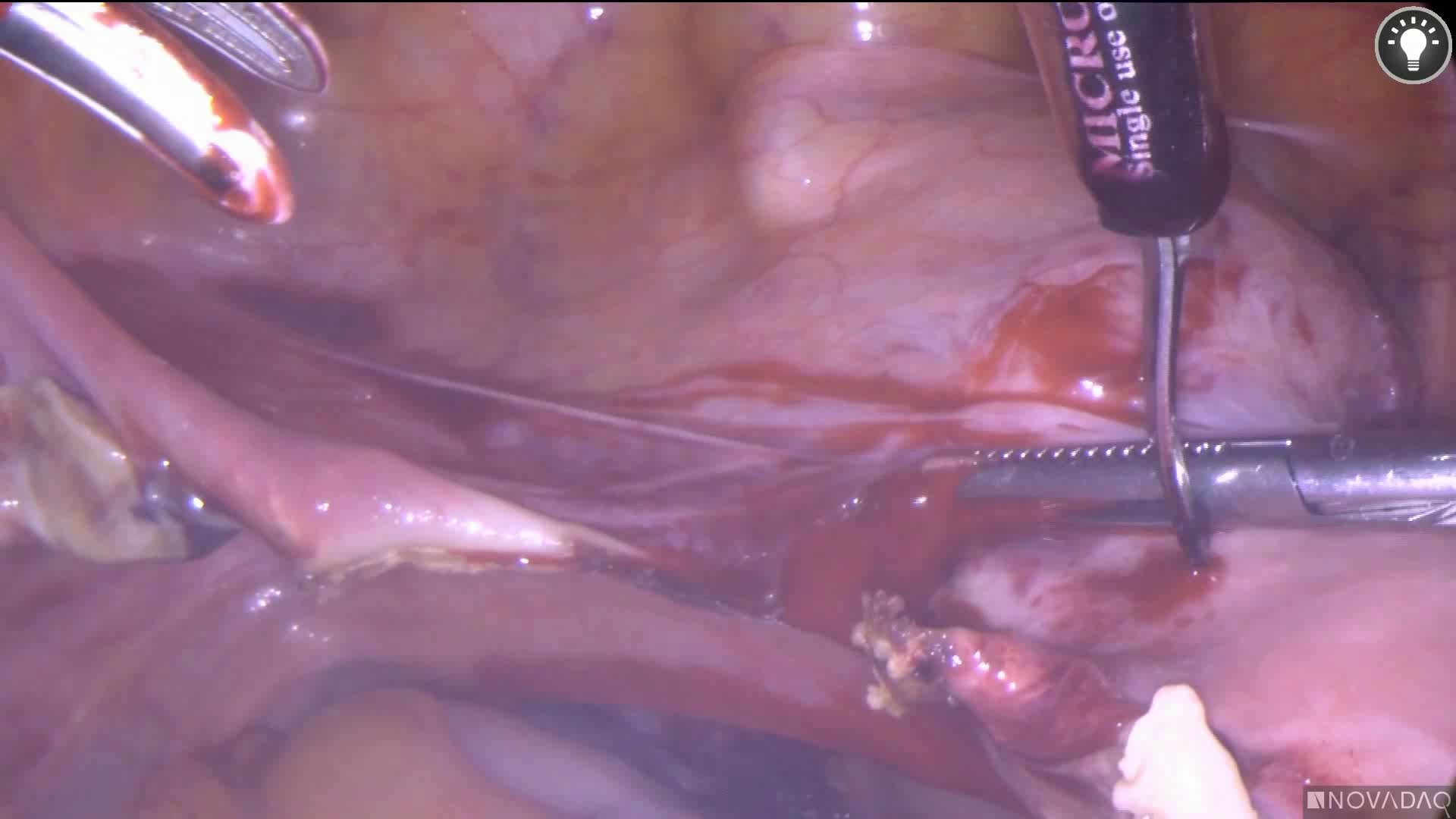}
    \caption{Bleeding}    
  \end{subfigure}
  \begin{subfigure}[t]{0.31\columnwidth}
    \centering
    \includegraphics[width=\linewidth]{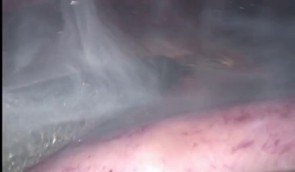}
    \caption{Smoke}    
  \end{subfigure}

  \caption{Sample frames illustrating different surgical actions and side-effects in gynecologic laparoscopy.}
  \Description{Sample frames illustrating different surgical actions and side-effects in gynecologic laparoscopy.}
  \label{fig:Exemplar_frames_action}
\end{figure}

In this work, we introduce GynSurg, a comprehensive multi-task dataset designed specifically for gynecologic laparoscopic surgery. GynSurg consists of 152 high-definition videos, expertly annotated with temporal labels for four critical surgical actions: (i) coagulation, (ii) needle passing, (iii) suction/irrigation, and (iv) transection, as well as two intraoperative side effects: (a) bleeding and (b) smoke. In addition, the dataset includes more than 12,000 frames with pixel-level segmentation masks covering all commonly used laparoscopic instruments and four key pelvic organs. By integrating temporal annotations for actions and side effects with spatial annotations for instruments and anatomy, GynSurg provides a unified resource that enables end-to-end evaluation across multiple tasks, including action recognition, side effect detection, and semantic segmentation of both surgical instruments and anatomical structures. A detailed comparison of GynSurg against existing gynecologic laparoscopy datasets in terms of task coverage, dataset scale, and annotation types is presented in Table~\ref{tab:Comparison of publicly dataset}.

The remainder of this paper is organized as follows. Section~\ref{sec:dataset} details our dataset -- its acquisition, annotation protocols, and key statistics. Section~\ref{sec:methods} describes the network architectures and training procedures used for both action recognition and semantic segmentation. In Section~\ref{sec:experimental results}, we present and analyze our experimental results. Finally, Section~\ref{sec:Conclusion} provides the conclusions from this study.

\section{Dataset}
\label{sec:dataset}
\begin{table}[b]
\centering
\caption{Timeline visualization of action and side-effect annotations in selected laparoscopic video cases.}
\label{tab:Timeline_visualization}
\resizebox{0.45\textwidth}{!}{%
\begin{tabular}{m{1.0cm}m{12cm}}
\specialrule{.12em}{.05em}{.05em}
Case & Actions \\\midrule
1&\DP{0.19}{Purple}\DP{0.35}{Gray}\DP{0.12}{Purple}\DP{1.09}{Gray}\DP{0.23}{Gray}\DP{0.19}{Gray}\DP{0.07}{Gray}\DP{-0.01}{Gray}\DP{0.01}{RoyalBlue}\DP{0.00}{Gray}\DP{0.01}{RoyalBlue}\DP{0.08}{Gray}\DP{0.04}{Gray}\DP{0.10}{Gray}\DP{0.03}{Gray}\DP{0.02}{Gray}\DP{0.00}{Gray}\DP{0.72}{Gray}\DP{0.03}{RoyalBlue}\DP{0.04}{Gray}\DP{0.00}{RoyalBlue}\DP{0.01}{Gray}\DP{0.04}{RoyalBlue}\DP{0.05}{Gray}\DP{0.01}{RoyalBlue}\DP{0.11}{Gray}\DP{0.19}{Purple}\DP{0.34}{Gray}\DP{0.08}{Purple}\DP{0.20}{Gray}\DP{0.06}{RoyalBlue}\DP{0.05}{Gray}\DP{0.11}{RoyalBlue}\DP{0.09}{Gray}\DP{0.07}{RoyalBlue}\DP{0.01}{Gray}\DP{0.05}{Gray}\DP{0.11}{Gray}\DP{0.06}{Gray}\DP{0.10}{Gray}\DP{0.03}{Gray}\DP{0.12}{Gray}\DP{0.03}{Gray}\DP{0.03}{Gray}\DP{0.00}{Gray}\DP{0.84}{Gray}\DP{0.27}{Purple}\DP{0.31}{Gray}\DP{0.07}{RoyalBlue}\DP{0.04}{Gray}\DP{0.13}{Gray}\DP{0.03}{Gray}\DP{0.04}{RoyalBlue}\DP{0.10}{Gray}\DP{0.05}{RoyalBlue}\DP{0.12}{Gray}\DP{0.02}{Gray}\DP{0.09}{Gray}\DP{0.01}{Gray}\DP{0.01}{Gray}\DP{0.04}{Gray}\DP{0.02}{Gray}\DP{0.00}{Gray}\DP{0.29}{Gray}\DP{0.01}{RoyalBlue}\DP{1.18}{Gray}\DP{0.07}{RoyalBlue}\DP{0.03}{Gray}\DP{0.03}{RoyalBlue}\DP{0.08}{Gray}\DP{0.03}{RoyalBlue}\DP{0.03}{Gray}\DP{0.01}{RoyalBlue}\DP{0.47}{Gray}\DP{0.13}{Purple}\DP{0.3}{Gray}\DP{0.02}{Purple}\\
2&\DP{1.01}{Gray}\DP{0.43}{Gray}\DP{0.15}{Purple}\DP{-0.05}{Gray}\DP{0.26}{Gray}\DP{0.62}{Gray}\DP{0.04}{Dandelion}\DP{0.27}{Gray}\DP{0.03}{RoyalBlue}\DP{0.17}{Gray}\DP{0.33}{Purple}\DP{0.17}{Gray}\DP{0.11}{Purple}\DP{0.89}{Gray}\DP{0.04}{Gray}\DP{0.13}{Gray}\DP{0.04}{Gray}\DP{0.11}{Gray}\DP{0.02}{Gray}\DP{0.19}{Gray}\DP{0.00}{Gray}\DP{0.89}{Gray}\DP{0.09}{Purple}\DP{0.25}{Gray}\DP{0.14}{Purple}\DP{0.34}{Gray}\DP{0.23}{Purple}\DP{0.16}{Gray}\DP{0.37}{Purple}\DP{0.60}{Gray}\DP{0.07}{Gray}\DP{0.12}{Gray}\DP{0.17}{Gray}\DP{0.01}{Gray}\DP{0.10}{RoyalBlue}\DP{0.00}{Gray}\DP{0.08}{RoyalBlue}\DP{0.54}{Gray}\DP{0.01}{Gray}\DP{0.77}{Gray}\DP{0.16}{Purple}\\
3&\DP{0.09}{Purple}\DP{0.15}{Gray}\DP{0.01}{RoyalBlue}\DP{0.02}{Gray}\DP{0.00}{RoyalBlue}\DP{0.01}{Gray}\DP{0.01}{RoyalBlue}\DP{0.02}{Gray}\DP{0.01}{RoyalBlue}\DP{0.05}{Gray}\DP{0.03}{RoyalBlue}\DP{0.06}{Gray}\DP{0.06}{RoyalBlue}\DP{0.01}{Gray}\DP{0.02}{RoyalBlue}\DP{0.00}{Gray}\DP{0.05}{RoyalBlue}\DP{0.04}{Gray}\DP{0.06}{Gray}\DP{0.12}{Gray}\DP{0.07}{Gray}\DP{0.08}{Gray}\DP{0.03}{Gray}\DP{0.12}{Gray}\DP{0.05}{Gray}\DP{0.07}{Gray}\DP{0.01}{Gray}\DP{0.53}{Gray}\DP{0.25}{Purple}\DP{0.44}{Gray}\DP{0.04}{Gray}\DP{0.13}{Gray}\DP{0.04}{Gray}\DP{0.17}{Gray}\DP{0.02}{Gray}\DP{0.25}{Gray}\DP{0.00}{Gray}\DP{0.38}{Gray}\DP{0.05}{RoyalBlue}\DP{0.00}{Gray}\DP{0.02}{RoyalBlue}\DP{0.02}{Gray}\DP{0.03}{RoyalBlue}\DP{0.02}{Gray}\DP{0.02}{RoyalBlue}\DP{0.09}{Gray}\DP{0.02}{RoyalBlue}\DP{0.01}{Gray}\DP{0.00}{RoyalBlue}\DP{0.47}{Gray}\DP{0.09}{Gray}\DP{0.02}{Gray}\DP{0.03}{RoyalBlue}\DP{0.11}{Gray}\DP{0.10}{Gray}\DP{-0.05}{Gray}\DP{0.01}{RoyalBlue}\DP{0.07}{Gray}\DP{0.08}{Gray}\DP{-0.03}{Gray}\DP{0.01}{RoyalBlue}\DP{0.10}{Gray}\DP{0.10}{Gray}\DP{-0.02}{Gray}\DP{0.01}{RoyalBlue}\DP{0.02}{Gray}\DP{0.01}{RoyalBlue}\DP{0.01}{Gray}\DP{0.01}{RoyalBlue}\DP{0.02}{Gray}\DP{0.06}{RoyalBlue}\DP{0.33}{Gray}\DP{0.01}{RoyalBlue}\DP{0.11}{Gray}\DP{0.09}{Gray}\DP{-0.01}{Gray}\DP{0.05}{RoyalBlue}\DP{0.02}{Gray}\DP{0.06}{RoyalBlue}\DP{0.14}{Gray}\DP{0.04}{Gray}\DP{0.04}{Gray}\DP{0.04}{RoyalBlue}\DP{0.01}{Gray}\DP{0.01}{RoyalBlue}\DP{0.03}{Gray}\DP{0.07}{RoyalBlue}\DP{0.99}{Gray}\DP{0.02}{RoyalBlue}\DP{0.18}{Gray}\DP{0.01}{RoyalBlue}\DP{0.03}{Gray}\DP{0.01}{RoyalBlue}\DP{0.03}{Gray}\DP{0.01}{RoyalBlue}\DP{0.02}{Gray}\DP{0.01}{RoyalBlue}\DP{0.02}{Gray}\DP{0.00}{RoyalBlue}\DP{0.01}{Gray}\DP{0.00}{RoyalBlue}\DP{0.02}{Gray}\DP{0.00}{RoyalBlue}\DP{0.01}{Gray}\DP{0.00}{RoyalBlue}\DP{0.02}{Gray}\DP{0.04}{RoyalBlue}\DP{0.18}{Gray}\DP{0.01}{RoyalBlue}\DP{0.01}{Gray}\DP{0.00}{RoyalBlue}\DP{0.02}{Gray}\DP{0.02}{RoyalBlue}\DP{2.19}{Gray}\DP{0.04}{Gray}\\
4&\DP{0.01}{Dandelion}\DP{0.20}{Gray}\DP{0.03}{Dandelion}\DP{0.06}{Gray}\DP{0.02}{Dandelion}\DP{0.11}{Gray}\DP{0.02}{Dandelion}\DP{0.01}{Gray}\DP{0.02}{Dandelion}\DP{0.10}{Gray}\DP{0.03}{Dandelion}\DP{0.02}{Gray}\DP{0.02}{Dandelion}\DP{0.06}{Gray}\DP{0.04}{Dandelion}\DP{0.03}{Gray}\DP{0.02}{Dandelion}\DP{0.03}{Gray}\DP{0.05}{Dandelion}\DP{0.03}{Gray}\DP{0.05}{Dandelion}\DP{0.03}{Gray}\DP{0.03}{Dandelion}\DP{0.19}{Gray}\DP{0.03}{Dandelion}\DP{0.02}{Gray}\DP{0.02}{Dandelion}\DP{0.03}{Gray}\DP{0.01}{Dandelion}\DP{2.03}{Gray}\DP{0.12}{Purple}\DP{0.33}{Gray}\DP{0.01}{Gray}\DP{0.37}{Gray}\DP{0.21}{Purple}\DP{0.42}{Gray}\DP{0.09}{Purple}\DP{0.14}{Gray}\DP{0.09}{Purple}\DP{0.33}{Gray}\DP{0.11}{Purple}\DP{0.15}{Gray}\DP{0.10}{Purple}\DP{0.11}{Gray}\DP{0.12}{Purple}\DP{0.22}{Gray}\DP{0.14}{Purple}\DP{0.27}{Gray}\DP{0.16}{Purple}\DP{0.09}{Gray}\DP{0.14}{Purple}\DP{0.17}{Gray}\DP{0.16}{Purple}\DP{0.28}{Gray}\DP{0.08}{Purple}\DP{0.28}{Gray}\DP{0.27}{Purple}\DP{0.10}{Gray}\DP{0.01}{Gray}\DP{0.09}{Gray}\DP{0.03}{RoyalBlue}\DP{0.00}{Gray}\DP{0.03}{RoyalBlue}\DP{0.00}{Gray}\DP{0.06}{RoyalBlue}\DP{0.02}{Gray}\DP{0.03}{RoyalBlue}\DP{0.00}{Gray}\DP{0.05}{RoyalBlue}\DP{0.00}{Gray}\DP{0.03}{RoyalBlue}\DP{0.00}{Gray}\DP{0.05}{RoyalBlue}\DP{0.00}{Gray}\DP{0.06}{RoyalBlue}\DP{0.01}{Gray}\DP{0.03}{RoyalBlue}\DP{0.00}{Gray}\DP{0.02}{RoyalBlue}\DP{0.02}{Gray}\DP{0.01}{Gray}\DP{0.01}{Gray}\DP{0.02}{RoyalBlue}\DP{0.02}{Gray}\DP{0.01}{RoyalBlue}\DP{0.00}{Gray}\DP{0.04}{RoyalBlue}\DP{0.05}{Gray}\DP{0.02}{RoyalBlue}\DP{0.11}{Gray}\DP{0.02}{RoyalBlue}\DP{0.00}{Gray}\DP{0.02}{RoyalBlue}\DP{0.00}{Gray}\DP{0.06}{RoyalBlue}\DP{0.00}{Gray}\DP{0.03}{RoyalBlue}\DP{0.00}{Gray}\DP{0.07}{RoyalBlue}\DP{0.01}{Gray}\DP{0.03}{RoyalBlue}\DP{0.25}{Gray}\DP{0.03}{RoyalBlue}\DP{0.00}{Gray}\DP{0.03}{RoyalBlue}\DP{0.30}{Gray}\DP{0.08}{RoyalBlue}\\
5&\DP{0.07}{ForestGreen}\DP{0.44}{Gray}\DP{0.10}{ForestGreen}\DP{0.16}{Gray}\DP{0.05}{ForestGreen}\DP{0.22}{Gray}\DP{0.07}{Gray}\DP{0.11}{Gray}\DP{0.25}{Gray}\DP{0.01}{Gray}\DP{0.03}{Gray}\DP{0.21}{Gray}\DP{0.01}{RoyalBlue}\DP{0.01}{Gray}\DP{0.08}{RoyalBlue}\DP{0.01}{Gray}\DP{0.04}{RoyalBlue}\DP{0.00}{Gray}\DP{0.05}{RoyalBlue}\DP{0.13}{Gray}\DP{0.04}{RoyalBlue}\DP{0.06}{Gray}\DP{0.02}{Dandelion}\DP{-0.01}{Gray}\DP{0.01}{Gray}\DP{0.02}{Gray}\DP{0.02}{RoyalBlue}\DP{0.00}{Gray}\DP{0.03}{RoyalBlue}\DP{0.03}{Gray}\DP{0.01}{RoyalBlue}\DP{0.00}{Gray}\DP{0.01}{RoyalBlue}\DP{0.01}{Gray}\DP{0.00}{RoyalBlue}\DP{0.04}{Gray}\DP{0.02}{RoyalBlue}\DP{0.06}{Gray}\DP{0.01}{RoyalBlue}\DP{0.04}{Gray}\DP{0.01}{Gray}\DP{0.02}{Gray}\DP{0.01}{RoyalBlue}\DP{0.15}{Gray}\DP{0.01}{Gray}\DP{0.02}{Gray}\DP{0.03}{Dandelion}\DP{-0.02}{Gray}\DP{0.03}{Gray}\DP{0.01}{Gray}\DP{0.01}{RoyalBlue}\DP{0.00}{Gray}\DP{0.03}{RoyalBlue}\DP{0.03}{Gray}\DP{0.03}{RoyalBlue}\DP{0.03}{Gray}\DP{0.03}{Dandelion}\DP{-0.02}{Gray}\DP{0.03}{Gray}\DP{0.00}{Gray}\DP{0.01}{RoyalBlue}\DP{0.03}{Gray}\DP{0.01}{RoyalBlue}\DP{0.01}{Gray}\DP{0.02}{RoyalBlue}\DP{0.02}{Gray}\DP{0.03}{Dandelion}\DP{-0.03}{Gray}\DP{0.04}{Gray}\DP{0.01}{Gray}\DP{0.01}{RoyalBlue}\DP{0.00}{Gray}\DP{0.01}{RoyalBlue}\DP{0.01}{Gray}\DP{0.03}{RoyalBlue}\DP{0.05}{Gray}\DP{0.02}{RoyalBlue}\DP{0.02}{Gray}\DP{0.01}{RoyalBlue}\DP{0.05}{Gray}\DP{0.01}{Dandelion}\DP{-0.01}{Gray}\DP{0.02}{Gray}\DP{0.03}{Gray}\DP{0.01}{RoyalBlue}\DP{0.00}{Gray}\DP{0.01}{RoyalBlue}\DP{0.02}{Gray}\DP{0.03}{RoyalBlue}\DP{0.07}{Gray}\DP{0.01}{RoyalBlue}\DP{0.00}{Gray}\DP{0.04}{RoyalBlue}\DP{0.02}{Gray}\DP{0.02}{RoyalBlue}\DP{0.01}{Gray}\DP{0.04}{RoyalBlue}\DP{0.01}{Gray}\DP{0.01}{RoyalBlue}\DP{0.30}{Gray}\DP{0.04}{Gray}\DP{0.86}{Gray}\DP{0.17}{Purple}\DP{0.18}{Gray}\DP{0.12}{Purple}\DP{0.27}{Gray}\DP{0.01}{Gray}\DP{0.23}{Gray}\DP{0.12}{Purple}\DP{0.15}{Gray}\DP{0.12}{Purple}\DP{0.15}{Gray}\DP{0.08}{Purple}\DP{0.15}{Gray}\DP{0.10}{Purple}\DP{0.16}{Gray}\DP{0.06}{Purple}\DP{0.23}{Gray}\DP{0.18}{Purple}\DP{0.09}{Gray}\DP{0.05}{Purple}\DP{0.21}{Gray}\DP{0.10}{Purple}\DP{0.22}{Gray}\DP{0.12}{Purple}\DP{0.14}{Gray}\DP{0.04}{Purple}\DP{0.19}{Gray}\DP{0.01}{Gray}\DP{0.18}{Gray}\DP{0.01}{RoyalBlue}\DP{0.02}{Gray}\DP{0.04}{RoyalBlue}\DP{0.00}{Gray}\DP{0.03}{RoyalBlue}\DP{0.01}{Gray}\DP{0.03}{RoyalBlue}\DP{0.03}{Gray}\DP{0.02}{RoyalBlue}\DP{0.01}{Gray}\DP{0.03}{RoyalBlue}\DP{0.04}{Gray}\DP{0.01}{RoyalBlue}\DP{0.05}{Gray}\DP{0.03}{RoyalBlue}\DP{0.00}{Gray}\DP{0.05}{RoyalBlue}\DP{0.04}{Gray}\DP{0.05}{RoyalBlue}\DP{0.06}{Gray}\DP{0.02}{RoyalBlue}\DP{0.17}{Gray}\DP{0.01}{Dandelion}\DP{-0.01}{Gray}\DP{0.01}{Gray}\DP{0.01}{Gray}\DP{0.04}{RoyalBlue}\DP{0.01}{Gray}\DP{0.01}{RoyalBlue}\DP{0.26}{Gray}\DP{0.01}{Dandelion}\DP{-0.01}{Gray}\DP{0.02}{Gray}\DP{0.03}{Gray}\DP{0.01}{Dandelion}\DP{0.01}{Gray}\DP{0.01}{RoyalBlue}\DP{0.02}{Gray}\DP{0.02}{Dandelion}\DP{-0.02}{Gray}\DP{0.03}{Gray}\DP{-0.01}{Gray}\DP{0.01}{RoyalBlue}\DP{0.00}{Gray}\DP{0.02}{RoyalBlue}\DP{0.04}{Gray}\DP{0.01}{Dandelion}\DP{-0.01}{Gray}\\
6&\DP{0.30}{Purple}\DP{0.24}{Gray}\DP{0.19}{Purple}\DP{0.20}{Gray}\DP{0.18}{Purple}\DP{1.22}{Gray}\DP{0.06}{Gray}\DP{0.13}{Gray}\DP{0.00}{Gray}\DP{0.20}{Gray}\DP{0.08}{RoyalBlue}\DP{0.18}{Gray}\DP{0.04}{RoyalBlue}\DP{0.00}{Gray}\DP{0.03}{RoyalBlue}\DP{0.01}{Gray}\DP{0.04}{RoyalBlue}\DP{0.03}{Gray}\DP{0.02}{RoyalBlue}\DP{0.00}{Gray}\DP{0.02}{RoyalBlue}\DP{0.26}{Gray}\DP{0.01}{RoyalBlue}\DP{0.01}{Gray}\DP{0.03}{RoyalBlue}\DP{-0.00}{Gray}\DP{0.04}{RoyalBlue}\DP{0.38}{Gray}\DP{0.20}{Purple}\DP{-0.00}{Gray}\DP{0.02}{Purple}\DP{0.11}{Gray}\DP{0.19}{Purple}\DP{0.26}{Gray}\DP{0.11}{Purple}\DP{0.12}{Gray}\DP{0.08}{Purple}\DP{1.29}{Gray}\DP{0.11}{Gray}\DP{0.03}{Gray}\DP{0.01}{Gray}\DP{0.28}{Gray}\DP{0.06}{RoyalBlue}\DP{0.00}{Gray}\DP{0.05}{RoyalBlue}\DP{0.02}{Gray}\DP{0.05}{RoyalBlue}\DP{0.01}{Gray}\DP{0.04}{RoyalBlue}\DP{0.16}{Gray}\DP{0.03}{RoyalBlue}\DP{0.00}{Gray}\DP{0.02}{RoyalBlue}\DP{0.07}{Gray}\DP{0.03}{RoyalBlue}\DP{0.00}{Gray}\DP{0.03}{RoyalBlue}\DP{0.14}{Gray}\DP{0.19}{Purple}\DP{0.11}{Gray}\DP{0.10}{Purple}\DP{0.02}{Gray}\DP{0.03}{RoyalBlue}\DP{0.00}{Gray}\DP{0.02}{RoyalBlue}\DP{0.19}{Gray}\DP{0.04}{RoyalBlue}\DP{0.00}{Gray}\DP{0.02}{RoyalBlue}\DP{0.07}{Gray}\DP{0.02}{RoyalBlue}\DP{0.00}{Gray}\DP{0.02}{RoyalBlue}\DP{0.52}{Gray}\DP{0.08}{Gray}\DP{0.03}{Gray}\DP{0.01}{Gray}\DP{0.62}{Gray}\DP{0.19}{ForestGreen}\DP{-0.18}{Gray}\DP{0.15}{Gray}\DP{0.1}{Gray}\DP{0.08}{ForestGreen}\\
7&\DP{0.11}{ForestGreen}\DP{0.23}{Gray}\DP{0.15}{ForestGreen}\DP{0.08}{Gray}\DP{0.18}{ForestGreen}\DP{0.05}{Gray}\DP{0.04}{ForestGreen}\DP{0.15}{Gray}\DP{0.18}{Dandelion}\DP{0.44}{Gray}\DP{0.03}{Dandelion}\DP{0.17}{Gray}\DP{0.04}{ForestGreen}\DP{0.16}{Gray}\DP{0.04}{ForestGreen}\DP{0.23}{Gray}\DP{0.03}{Gray}\DP{0.05}{Gray}\DP{0.02}{Gray}\DP{0.09}{Gray}\DP{0.08}{ForestGreen}\DP{0.14}{Gray}\DP{0.02}{ForestGreen}\DP{0.68}{Gray}\DP{0.05}{ForestGreen}\DP{0.13}{Gray}\DP{0.13}{ForestGreen}\DP{0.41}{Gray}\DP{0.08}{ForestGreen}\DP{0.20}{Gray}\DP{0.03}{ForestGreen}\DP{0.31}{Gray}\DP{0.03}{ForestGreen}\DP{0.03}{Gray}\DP{0.11}{ForestGreen}\DP{0.67}{Gray}\DP{0.24}{ForestGreen}\DP{0.12}{Gray}\DP{0.07}{ForestGreen}\DP{1.20}{Gray}\DP{0.09}{RoyalBlue}\DP{0.08}{Gray}\DP{0.03}{RoyalBlue}\DP{0.19}{Gray}\DP{0.04}{RoyalBlue}\DP{-0.02}{Gray}\DP{0.01}{Gray}\DP{0.24}{Gray}\DP{0.03}{RoyalBlue}\DP{0.04}{Gray}\DP{0.01}{RoyalBlue}\DP{0.10}{Gray}\DP{0.02}{RoyalBlue}\DP{0.01}{Gray}\DP{0.02}{RoyalBlue}\DP{0.17}{Gray}\DP{0.37}{Dandelion}\DP{0.07}{Gray}\DP{0.19}{Dandelion}\DP{0.22}{Gray}\DP{0.12}{Dandelion}\DP{0.03}{Gray}\DP{0.03}{Dandelion}\DP{0.03}{Gray}\DP{0.06}{Dandelion}\DP{0.14}{Gray}\DP{0.03}{RoyalBlue}\DP{0.06}{Gray}\DP{0.03}{Gray}\DP{0.19}{Gray}\DP{0.02}{Gray}\DP{0.03}{Gray}\DP{0.25}{Dandelion}\\
8&\DP{0.46}{Gray}\DP{0.00}{Gray}\DP{0.51}{Dandelion}\DP{1.39}{Gray}\DP{0.04}{Gray}\DP{1.69}{Gray}\DP{0.35}{Purple}\DP{0.26}{Gray}\DP{0.02}{Gray}\DP{0.55}{Gray}\DP{0.06}{Purple}\DP{0.41}{Gray}\DP{0.19}{Purple}\DP{0.30}{Gray}\DP{0.23}{Purple}\DP{0.45}{Gray}\DP{0.17}{Purple}\DP{0.08}{Gray}\DP{0.06}{Purple}\DP{0.53}{Gray}\DP{0.35}{Purple}\DP{0.39}{Gray}\DP{0.30}{Purple}\DP{0.48}{Gray}\DP{0.01}{Gray}\DP{0.23}{Gray}\DP{0.10}{RoyalBlue}\DP{0.09}{Gray}\DP{0.04}{RoyalBlue}\DP{0.17}{Gray}\DP{0.02}{RoyalBlue}\DP{-0.01}{Gray}\DP{0.02}{RoyalBlue}\DP{0.00}{Gray}\DP{0.07}{RoyalBlue}\DP{-0.07}{Gray}\DP{0.06}{RoyalBlue}\\
9&\DP{0.58}{Dandelion}\DP{0.18}{Gray}\DP{0.06}{RoyalBlue}\DP{0.03}{Gray}\DP{0.39}{Gray}\DP{-0.32}{Gray}\DP{0.44}{Dandelion}\DP{0.26}{Gray}\DP{0.05}{Dandelion}\DP{0.34}{Gray}\DP{0.05}{Dandelion}\DP{0.46}{Gray}\DP{0.18}{Dandelion}\DP{0.10}{Gray}\DP{0.09}{Dandelion}\DP{0.15}{Gray}\DP{0.11}{Dandelion}\DP{0.21}{Gray}\DP{0.08}{Dandelion}\DP{0.33}{Gray}\DP{0.08}{ForestGreen}\DP{0.94}{Gray}\DP{1.12}{Gray}\DP{0.15}{ForestGreen}\DP{0.73}{Gray}\DP{0.21}{ForestGreen}\DP{0.39}{Gray}\DP{0.13}{ForestGreen}\DP{0.12}{Gray}\DP{0.20}{ForestGreen}\DP{0.04}{Gray}\DP{0.06}{Gray}\DP{0.14}{Gray}\DP{0.08}{Gray}\DP{-0.07}{Gray}\DP{0.21}{Dandelion}\DP{0.24}{Gray}\DP{0.10}{ForestGreen}\DP{0.24}{Gray}\DP{0.10}{ForestGreen}\DP{0.06}{Gray}\DP{0.20}{ForestGreen}\DP{0.09}{Gray}\DP{0.13}{ForestGreen}\\
10&\DP{0.09}{ForestGreen}\DP{-0.09}{Gray}\DP{0.10}{Gray}\DP{0.02}{Gray}\DP{0.16}{ForestGreen}\DP{0.06}{Gray}\DP{0.15}{ForestGreen}\DP{-0.08}{Gray}\DP{0.00}{Gray}\DP{0.04}{Gray}\DP{0.00}{Gray}\DP{0.05}{Gray}\DP{0.02}{Gray}\DP{0.15}{Gray}\DP{0.05}{Gray}\DP{-0.01}{Gray}\DP{0.01}{Gray}\DP{8.0}{Gray}\DP{0.01}{RoyalBlue}\DP{0.00}{Gray}\DP{0.04}{RoyalBlue}\DP{0.00}{Gray}\DP{0.03}{RoyalBlue}\DP{0.01}{Gray}\DP{0.08}{RoyalBlue}\DP{0.04}{Gray}\DP{0.00}{RoyalBlue}\DP{0.02}{Gray}\DP{0.03}{RoyalBlue}\DP{0.05}{Gray}\DP{0.03}{RoyalBlue}\DP{0.11}{Gray}\DP{0.04}{RoyalBlue}\DP{0.18}{Gray}\DP{0.04}{RoyalBlue}\DP{0.00}{Gray}\DP{0.02}{RoyalBlue}\DP{0.36}{Gray}\DP{0.02}{RoyalBlue}\DP{0.01}{Gray}\DP{0.01}{RoyalBlue}\\
\midrule
\end{tabular}
}
\resizebox{0.45\textwidth}{!}{%
\begin{tabular}{m{1.5cm}m{13 cm}}
Colormap &  Coagulation \DPlegend{0.9}{Dandelion}, 
NeedlePassing \DPlegend{0.9}{Purple},
Suction/Irrigation \DPlegend{0.9}{RoyalBlue},
Transection \DPlegend{0.9}{ForestGreen},
Rest \DPlegend{0.9}{Gray}
\\
\specialrule{.12em}{.05em}{.05em}
\end{tabular}
}
\resizebox{0.45\textwidth}{!}{%
\begin{tabular}{m{1.0cm}m{12cm}}
Case & Bleeding \\\midrule
1&\DP{0.01}{Gray}\DP{0.04}{Gray}\DP{0.01}{Gray}\DP{0.39}{Gray}\DP{0.07}{Gray}\DP{0.09}{Gray}\DP{0.00}{Gray}\DP{0.08}{Gray}\DP{0.00}{Gray}\DP{0.03}{Gray}\DP{0.02}{Gray}\DP{0.05}{Gray}\DP{0.04}{Gray}\DP{0.07}{Gray}\DP{0.00}{Gray}\DP{0.01}{Gray}\DP{0.03}{Maroon}\DP{0.00}{Gray}\DP{0.07}{Gray}\DP{0.21}{Gray}\DP{0.00}{Gray}\DP{0.05}{Gray}\DP{0.00}{Gray}\DP{0.01}{Gray}\DP{0.00}{Gray}\DP{0.02}{Gray}\DP{0.00}{Gray}\DP{0.01}{Gray}\DP{0.01}{Gray}\DP{0.03}{Gray}\DP{0.13}{Gray}\DP{0.02}{Gray}\DP{0.00}{Gray}\DP{0.02}{Gray}\DP{0.03}{Gray}\DP{0.02}{Gray}\DP{0.08}{Gray}\DP{0.47}{Gray}\DP{0.15}{Gray}\DP{0.02}{Gray}\DP{0.05}{Gray}\DP{0.11}{Gray}\DP{0.06}{Maroon}\DP{0.44}{Gray}\DP{0.38}{Gray}\DP{0.17}{Gray}\DP{0.04}{Maroon}\DP{0.05}{Gray}\DP{0.01}{Maroon}\DP{0.04}{Gray}\DP{0.08}{Maroon}\DP{0.00}{Gray}\DP{0.02}{Gray}\DP{0.04}{Gray}\DP{0.01}{Gray}\DP{0.62}{Gray}\DP{0.27}{Gray}\DP{0.02}{Gray}\DP{0.09}{Gray}\DP{0.08}{Gray}\DP{0.12}{Gray}\DP{0.45}{Gray}\DP{0.08}{Maroon}\DP{0.65}{Gray}\DP{0.02}{Maroon}\DP{0.00}{Gray}\DP{0.01}{Gray}\DP{0.01}{Gray}\DP{0.03}{Gray}\DP{0.13}{Gray}\DP{0.17}{Maroon}\DP{0.29}{Gray}\DP{0.07}{Gray}\DP{1.71}{Gray}\DP{0.00}{Gray}\DP{0.31}{Gray}\DP{0.08}{Gray}\DP{0.09}{Gray}\DP{0.11}{Gray}\DP{0.08}{Gray}\DP{0.01}{Gray}\DP{0.25}{Gray}\DP{0.10}{Gray}\DP{0.11}{Gray}\DP{0.04}{Maroon}\DP{0.12}{Gray}\DP{0.1}{Gray}\\
2&\DP{0.09}{Gray}\DP{0.06}{Gray}\DP{0.09}{Gray}\DP{0.03}{Gray}\DP{0.08}{Gray}\DP{0.08}{Gray}\DP{0.07}{Gray}\DP{0.05}{Gray}\DP{0.07}{Gray}\DP{0.08}{Gray}\DP{0.07}{Gray}\DP{0.07}{Gray}\DP{0.07}{Maroon}\DP{0.07}{Gray}\DP{0.05}{Gray}\DP{0.08}{Gray}\DP{0.07}{Gray}\DP{0.11}{Gray}\DP{0.09}{Gray}\DP{0.05}{Gray}\DP{0.08}{Gray}\DP{0.22}{Gray}\DP{0.10}{Gray}\DP{0.00}{Gray}\DP{0.09}{Gray}\DP{0.08}{Gray}\DP{0.19}{Maroon}\DP{0.05}{Gray}\DP{0.13}{Maroon}\DP{0.07}{Gray}\DP{0.07}{Gray}\DP{0.14}{Gray}\DP{1.14}{Maroon}\DP{0.20}{Gray}\DP{0.07}{Gray}\DP{0.09}{Gray}\DP{0.06}{Gray}\DP{0.06}{Gray}\DP{0.06}{Gray}\DP{0.09}{Gray}\DP{0.09}{Gray}\DP{0.00}{Gray}\DP{1.08}{Maroon}\DP{0.39}{Gray}\DP{0.37}{Maroon}\DP{0.48}{Gray}\DP{0.13}{Gray}\DP{0.15}{Gray}\DP{0.32}{Gray}\DP{0.29}{Gray}\DP{0.44}{Gray}\DP{0.14}{Gray}\DP{0.17}{Gray}\DP{0.06}{Gray}\DP{0.26}{Gray}\DP{0.35}{Gray}\DP{0.29}{Gray}\DP{0.36}{Gray}\DP{0.2}{Gray}\\
3&\DP{0.39}{Gray}\DP{-0.01}{Gray}\DP{0.12}{Maroon}\DP{0.71}{Gray}\DP{0.35}{Maroon}\DP{0.11}{Gray}\DP{0.15}{Maroon}\DP{0.31}{Gray}\DP{0.44}{Maroon}\DP{1.10}{Gray}\DP{0.47}{Gray}\DP{0.30}{Gray}\DP{0.11}{Maroon}\DP{0.19}{Gray}\DP{0.17}{Maroon}\DP{0.44}{Gray}\DP{0.07}{Maroon}\DP{1.07}{Gray}\DP{0.51}{Gray}\DP{0.92}{Gray}\DP{0.98}{Maroon}\DP{0.33}{Gray}\DP{0.59}{Maroon}\DP{-0.20}{Gray}\\

\midrule
\end{tabular}
}
\resizebox{0.45\textwidth}{!}{%
\begin{tabular}{m{1.5cm}m{13 cm}}
Colormap &  Bleeding \DPlegend{0.9}{Maroon}, 
Non-bleeding \DPlegend{0.9}{Gray}
\\
\specialrule{.12em}{.05em}{.05em}
\end{tabular}
}
\resizebox{0.45\textwidth}{!}{%
\begin{tabular}{m{1.0cm}m{12cm}}
Case & Smoke \\\midrule
1&\DP{0.63}{YellowGreen}\DP{0.06}{Gray}\DP{0.06}{YellowGreen}\DP{0.31}{Gray}\DP{0.12}{YellowGreen}\DP{0.04}{Gray}\DP{0.26}{YellowGreen}\DP{0.12}{Gray}\DP{0.07}{YellowGreen}\DP{0.46}{Gray}\DP{0.24}{YellowGreen}\DP{0.10}{Gray}\DP{0.01}{YellowGreen}\DP{0.02}{Gray}\DP{0.08}{YellowGreen}\DP{0.11}{Gray}\DP{0.18}{YellowGreen}\DP{-0.11}{Gray}\DP{0.03}{Gray}\DP{0.14}{Gray}\DP{0.06}{YellowGreen}\DP{2.33}{Gray}\DP{1.50}{YellowGreen}\DP{0.08}{Gray}\DP{0.01}{YellowGreen}\DP{0.07}{Gray}\DP{0.36}{YellowGreen}\DP{0.10}{Gray}\DP{2.27}{YellowGreen}\\
2&\DP{1.13}{YellowGreen}\DP{0.42}{Gray}\DP{0.28}{YellowGreen}\DP{0.08}{Gray}\DP{0.26}{YellowGreen}\DP{0.81}{Gray}\DP{0.09}{YellowGreen}\DP{0.05}{Gray}\DP{0.09}{YellowGreen}\DP{0.30}{Gray}\DP{0.23}{YellowGreen}\DP{0.40}{Gray}\DP{0.24}{YellowGreen}\DP{0.39}{Gray}\DP{0.44}{YellowGreen}\DP{0.47}{Gray}\DP{0.54}{YellowGreen}\DP{0.25}{Gray}\DP{0.24}{YellowGreen}\DP{1.42}{Gray}\DP{1.07}{YellowGreen}\DP{0.71}{Gray}\DP{0.08}{YellowGreen}\\
3&\DP{0.20}{YellowGreen}\DP{-0.11}{Gray}\DP{0.18}{Gray}\DP{0.04}{Gray}\DP{0.06}{Gray}\DP{0.15}{Gray}\DP{0.10}{Gray}\DP{0.38}{Gray}\DP{0.19}{Gray}\DP{-0.05}{Gray}\DP{0.26}{Gray}\DP{-0.15}{Gray}\DP{0.15}{YellowGreen}\DP{0.26}{Gray}\DP{0.22}{YellowGreen}\DP{0.21}{Gray}\DP{0.09}{YellowGreen}\DP{0.12}{Gray}\DP{0.25}{YellowGreen}\DP{0.11}{Gray}\DP{0.12}{YellowGreen}\DP{0.08}{Gray}\DP{0.15}{YellowGreen}\DP{0.28}{Gray}\DP{0.25}{YellowGreen}\DP{0.02}{Gray}\DP{0.12}{YellowGreen}\DP{0.44}{Gray}\DP{0.19}{YellowGreen}\DP{2.30}{Gray}\DP{0.75}{YellowGreen}\DP{0.18}{Gray}\DP{0.32}{YellowGreen}\DP{0.62}{Gray}\DP{0.09}{YellowGreen}\DP{0.04}{Gray}\DP{0.24}{YellowGreen}\DP{0.08}{Gray}\DP{0.08}{YellowGreen}\DP{0.26}{Gray}\DP{0.08}{Gray}\DP{0.11}{YellowGreen}\\
\midrule
\end{tabular}
}
\resizebox{0.45\textwidth}{!}{%
\begin{tabular}{m{1.5cm}m{13 cm}}
Colormap &  Smoke \DPlegend{0.9}{YellowGreen}, 
Non-smoke \DPlegend{0.9}{Gray}
\\
\specialrule{.12em}{.05em}{.05em}
\end{tabular}
}
\end{table}

We introduce \textbf{GynSurg}, the first comprehensive multi-task dataset for gynecologic laparoscopic surgery, designed to support a wide range of video-based surgical analysis tasks. GynSurg integrates high-resolution video, dense temporal annotations for surgical actions and side effects, and pixel-level segmentation masks for surgical instruments and anatomical structures. This unified dataset provides a foundation for advancing research in surgical action recognition, instrument and anatomical segmentation, intraoperative event detection, and workflow understanding. The structure and contents of each dataset subset are detailed in the following subsections.

\begin{figure*}[t!]
    \centering
    \begin{subfigure}[t]{0.55\textwidth}
        \centering
        \includegraphics[width=\textwidth]{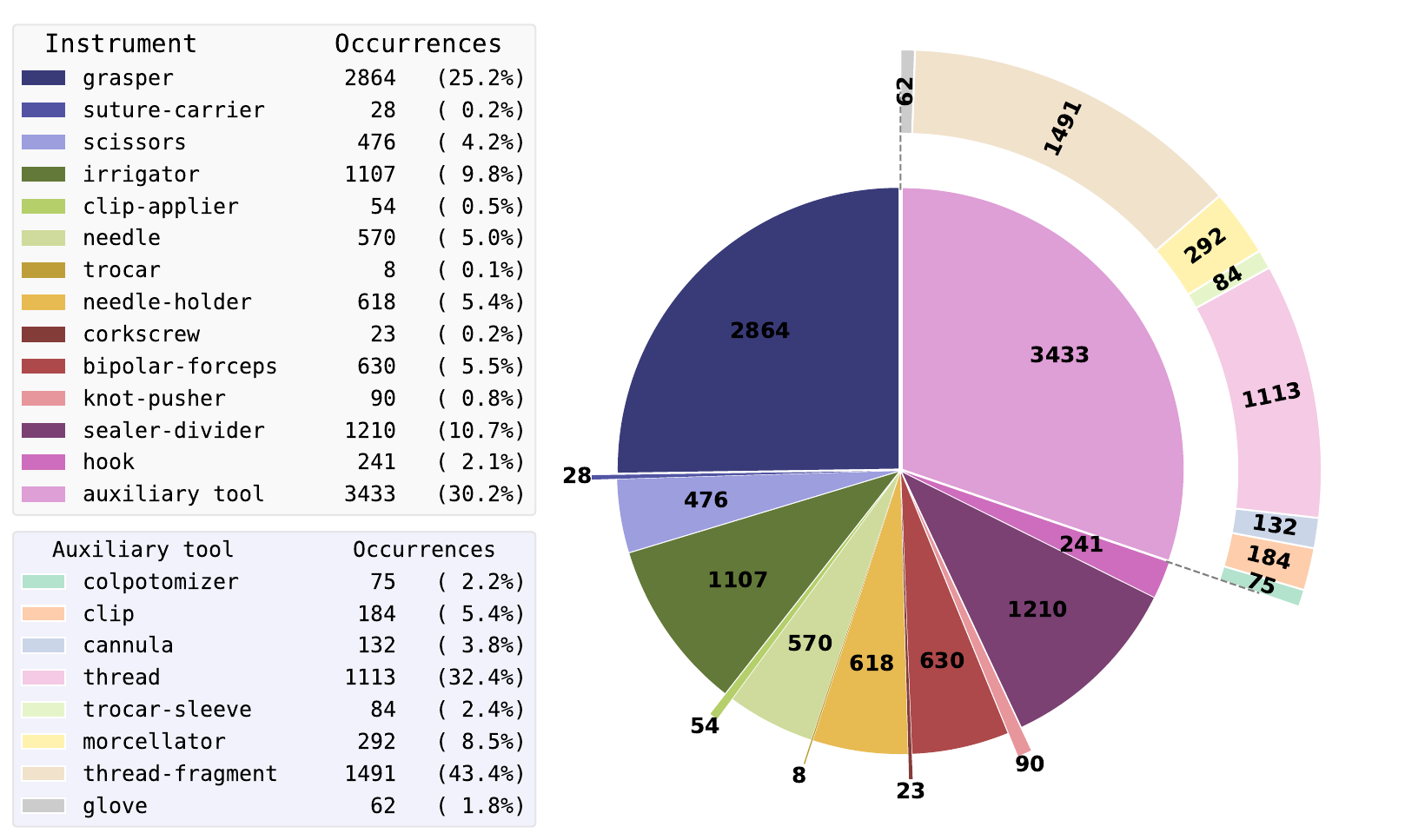}
        \caption{Instrument instance counts.}
        \label{fig:instrument-pie}
    \end{subfigure}
    \hfill
    \begin{subfigure}[t]{0.43\textwidth}
        \centering
        \includegraphics[width=\textwidth]{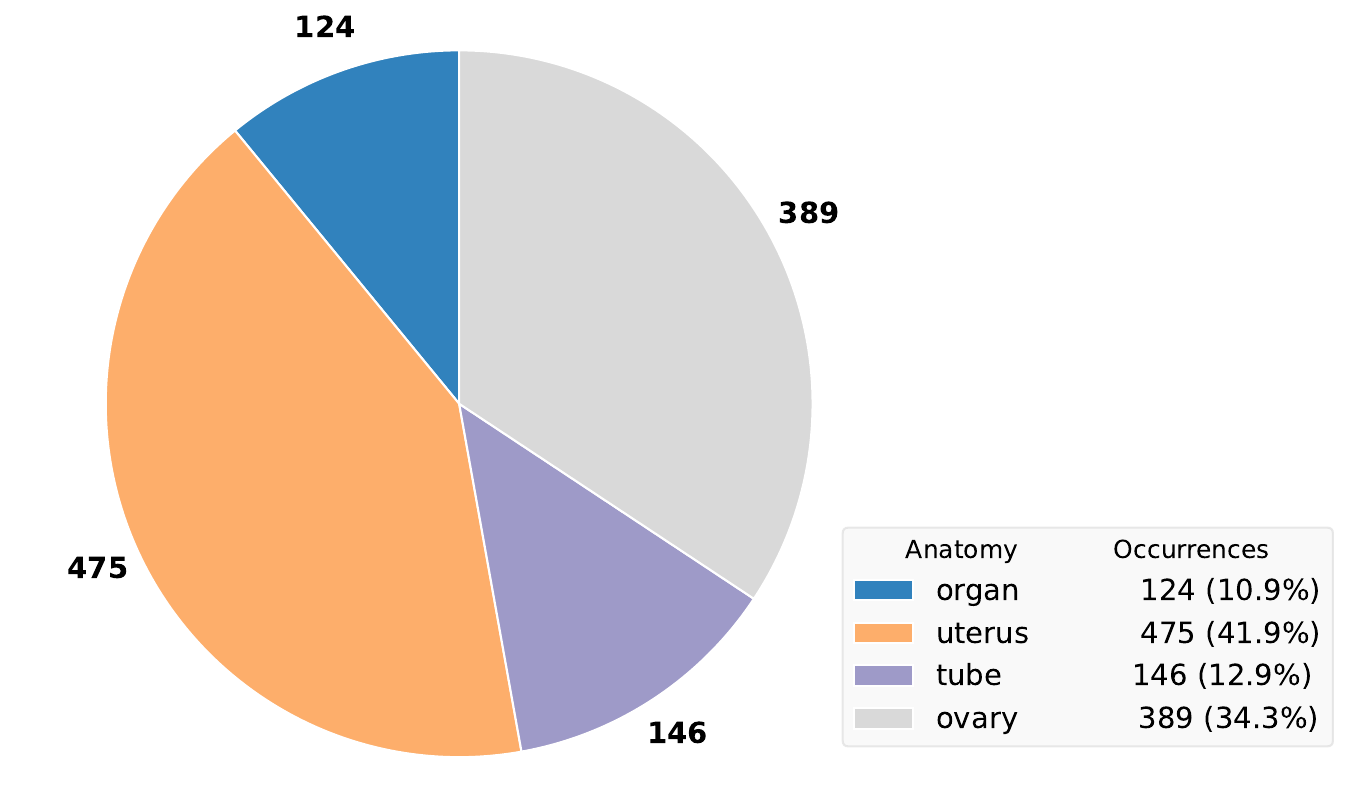}
        \caption{Anatomy instance counts.}
        \label{fig:anatomy-pie}
    \end{subfigure}
    \caption{Statistics of instrument (left) and anatomical structures (right) instances in the laparoscopy dataset.}
    \Description{Statistics of instrument (left) and anatomical structures (right) instances in the laparoscopy dataset.}
    \label{fig:Statistics_instrument_and_anatomies}
\end{figure*}

\subsection{Action Recognition Dataset}
\label{sec:action-recognition}

The action recognition component of GynSurg consists of 152 gynecologic laparoscopic surgery videos, selected from over 600 recorded procedures at the Medical University of Vienna and Medical University of Toronto. All videos were captured at 30 frames per second (fps) with a resolution of $1920 \times 1080$ pixels. Each video was meticulously annotated by clinical experts for four key operative actions: \textit{coagulation}, \textit{needle passing}, \textit{suction/irrigation}, and \textit{transection}. Additionally, a dedicated subset of the data was annotated for two intraoperative side effects: \textit{bleeding} and \textit{smoke}, supporting research on complex action and event recognition in laparoscopic surgery.

Table~\ref{tab:dataset_composition} summarizes the dataset composition across procedural actions and side effect labels. In addition to the primary laparoscopic actions, a “Rest” category is defined to represent intervals with no instrument activity or irrelevant motion. Side effects are annotated as binary classification tasks (\textit{bleeding} vs. \textit{non-bleeding}, \textit{smoke} vs. \textit{non-smoke}). The “Segments” column reports the number of extracted video segments per class, while the “Total Duration” column gives the cumulative length of all segments. To standardize the data for model development, video segments are further partitioned into three-second clips (with a one-second overlap), with the resulting clip counts also reported. Figure~\ref{fig:Exemplar_frames_action} presents example frames illustrating each surgical action and side effect.

\begin{figure*}[t!]
  \centering
  \begin{subfigure}[t]{0.16\textwidth}
    \centering
    \includegraphics[width=1\textwidth]{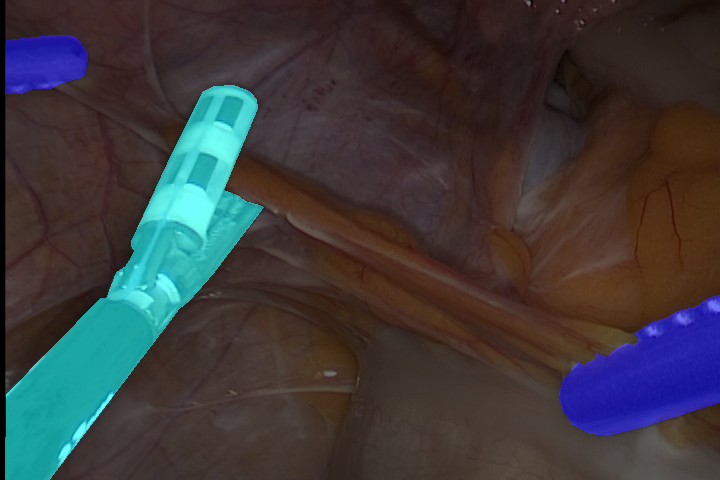}
  \end{subfigure}
  \begin{subfigure}[t]{0.16\textwidth}
    \centering
    \includegraphics[width=1\textwidth]{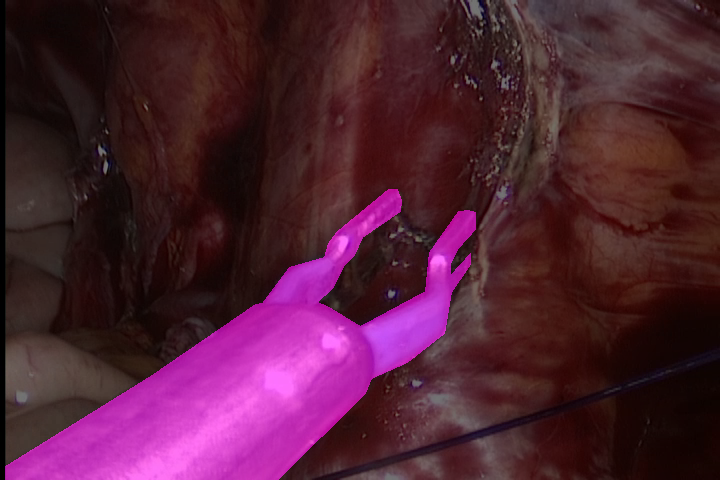}
  \end{subfigure}
  \begin{subfigure}[t]{0.16\textwidth}
    \centering
    \includegraphics[width=1\textwidth]{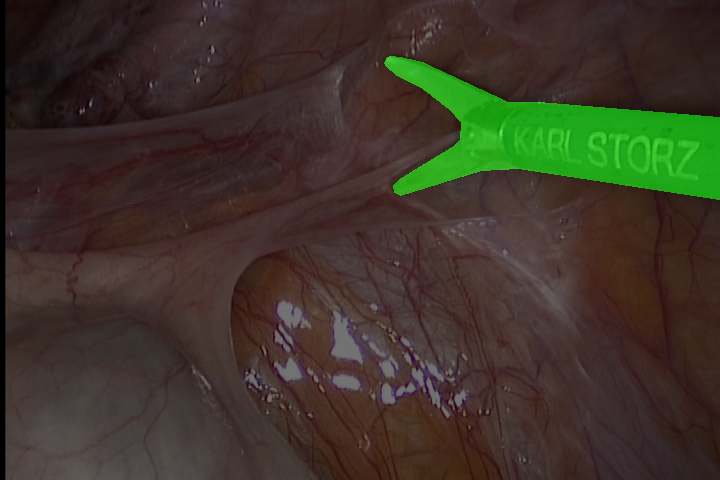}
  \end{subfigure}
  \begin{subfigure}[t]{0.16\textwidth}
    \centering
    \includegraphics[width=1\textwidth]{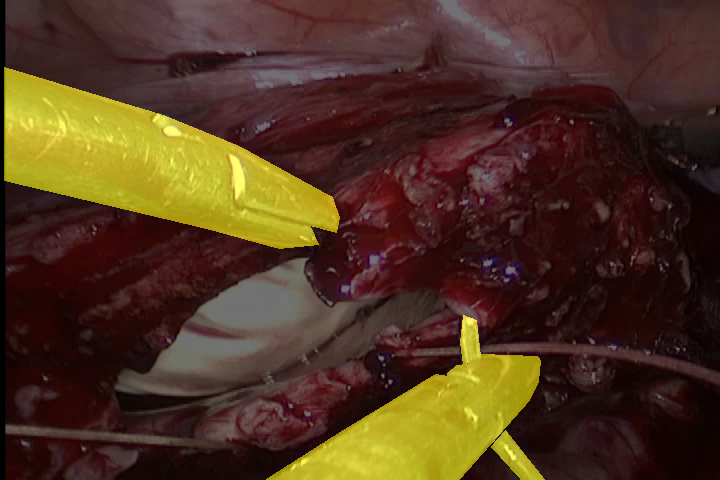}
  \end{subfigure}
  \begin{subfigure}[t]{0.16\textwidth}
    \centering
    \includegraphics[width=1\textwidth]{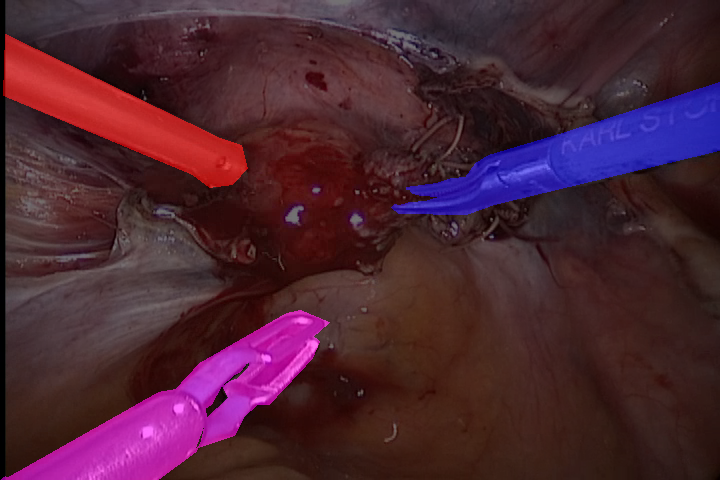}
  \end{subfigure}
  \begin{subfigure}[t]{0.16\textwidth}
    \centering
    \includegraphics[width=1\textwidth]{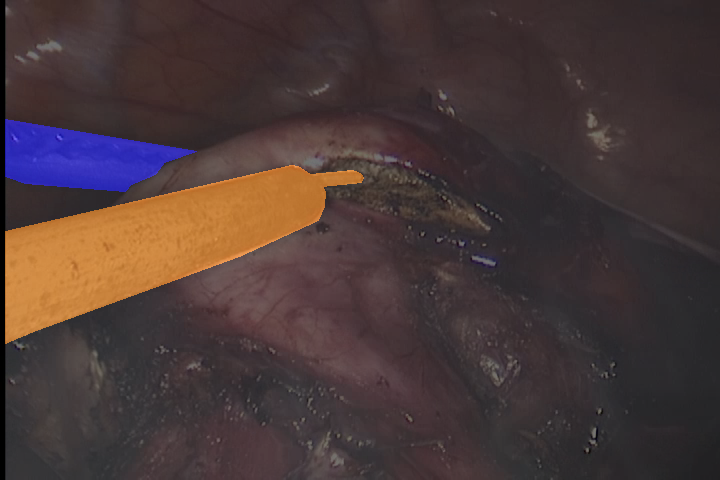}
  \end{subfigure}
  

  \begin{subfigure}[t]{0.16\textwidth}
    \centering
    \includegraphics[width=1\textwidth]{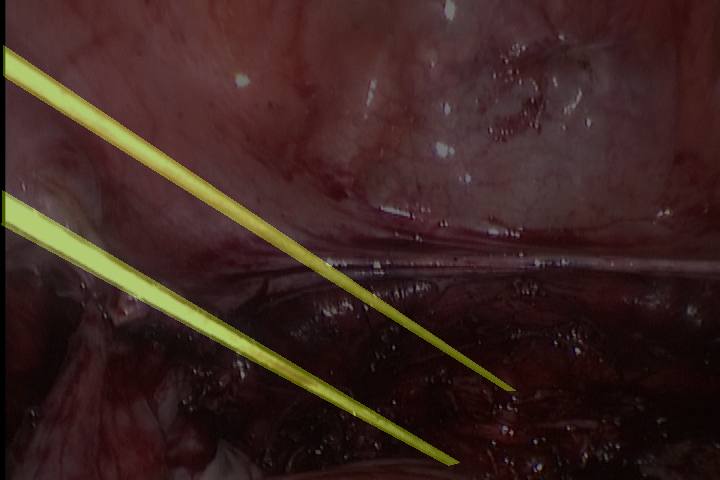}
  \end{subfigure}
  \begin{subfigure}[t]{0.16\textwidth}
    \centering
    \includegraphics[width=1\textwidth]{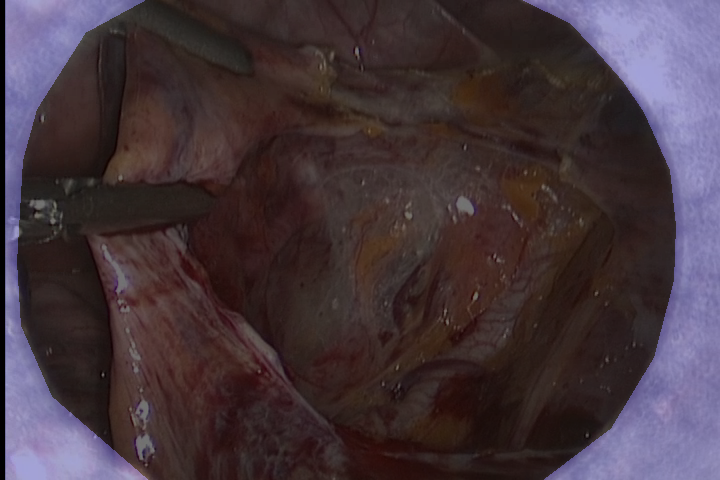}
  \end{subfigure}
  \begin{subfigure}[t]{0.16\textwidth}
    \centering
    \includegraphics[width=1\textwidth]{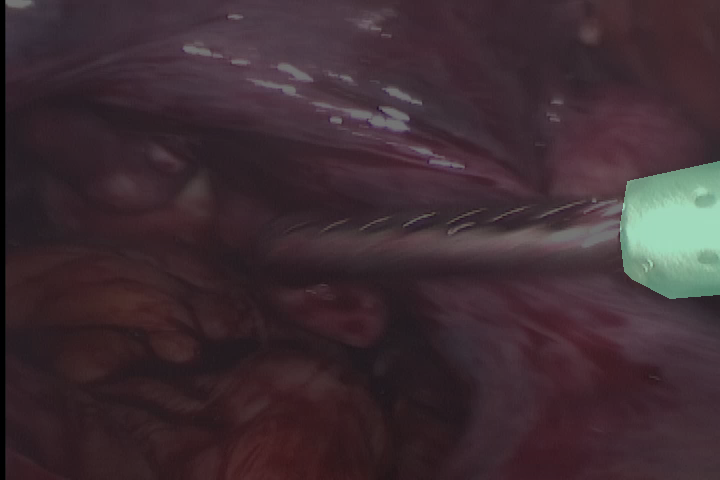}
  \end{subfigure}
  \begin{subfigure}[t]{0.16\textwidth}
    \centering
    \includegraphics[width=1\textwidth]{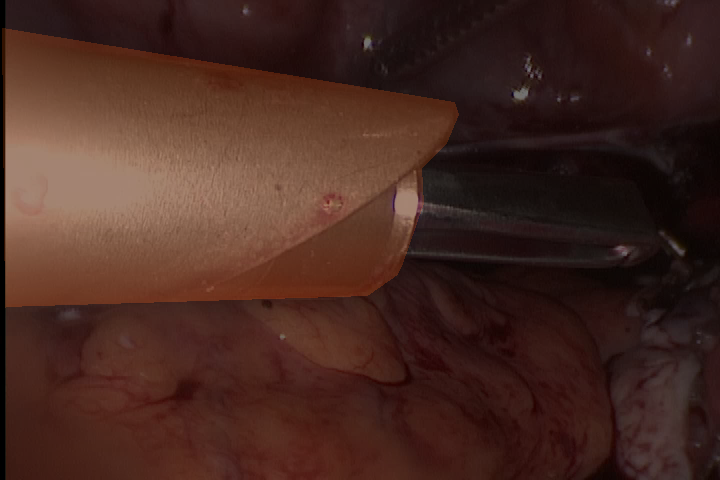}
  \end{subfigure}
  \begin{subfigure}[t]{0.16\textwidth}
    \centering
    \includegraphics[width=1\textwidth]{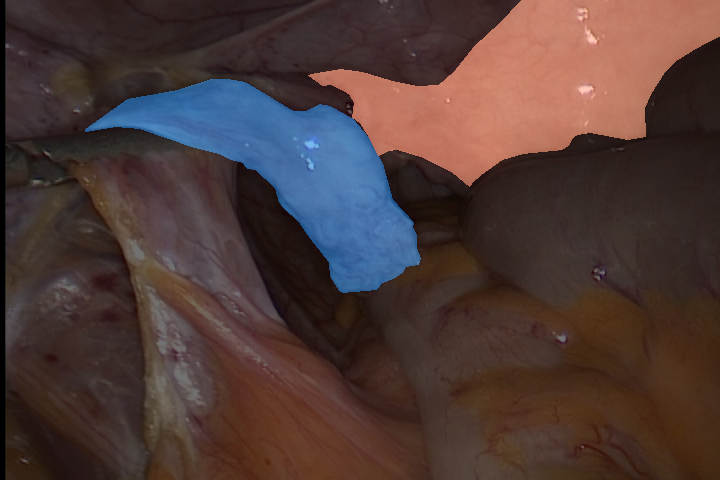}
  \end{subfigure}
  \begin{subfigure}[t]{0.16\textwidth}
    \centering
    \includegraphics[width=1\textwidth]{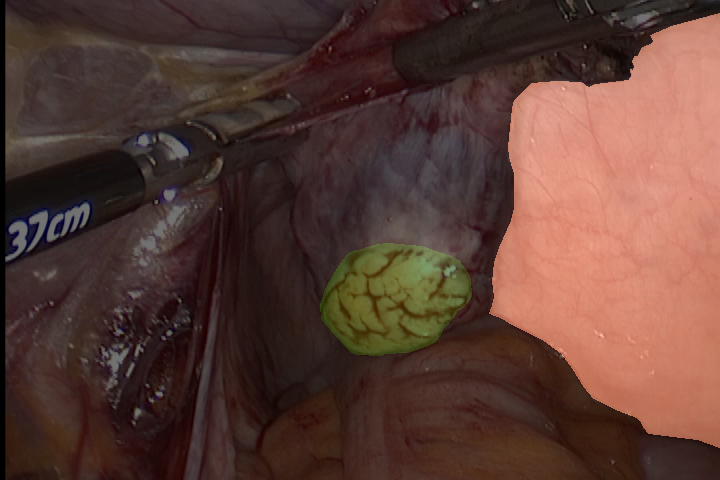}
  \end{subfigure}

  \caption{Visualization of pixel‐level annotations for surgical instruments and anatomical structures in the GynSurg dataset. Each instrument and anatomical structure is color-coded as follows: 
  grasper: \DP{0.9}{grasper}, 
  sealer-divider: \DP{0.9}{sealerdivider}, 
  bipolar forceps: \DP{0.9}{bipolarforceps},
  scissors: \DP{0.9}{scissors},
  suturing instrument: \DP{0.9}{suturinginstrument}, 
  irrigator: \DP{0.9}{irrigator}, 
  hook: \DP{0.9}{hook},
  thread: \DP{0.9}{thread},
  cannula: \DP{0.9}{cannula},
  trocar sleeve: \DP{0.9}{trocarsleeve}, 
  morcellator: \DP{0.9}{morcellator},  
  uterus: \DP{0.9}{uterus}, 
  tube: \DP{0.9}{tube}, 
  ovary: \DP{0.9}{ovary}).}
  \Description{Visualization of pixel‐level annotations for surgical instruments and anatomical structures in the GynSurg dataset. Each instrument and anatomical structure is color-coded as follows: 
  grasper: \DP{0.9}{grasper}, 
  sealer-divider: \DP{0.9}{sealerdivider}, 
  bipolar forceps: \DP{0.9}{bipolarforceps},
  scissors: \DP{0.9}{scissors},
  suturing instrument: \DP{0.9}{suturinginstrument}, 
  irrigator: \DP{0.9}{irrigator}, 
  hook: \DP{0.9}{hook},
  thread: \DP{0.9}{thread},
  cannula: \DP{0.9}{cannula},
  trocar sleeve: \DP{0.9}{trocarsleeve}, 
  morcellator: \DP{0.9}{morcellator},  
  uterus: \DP{0.9}{uterus}, 
  tube: \DP{0.9}{tube}, 
  ovary: \DP{0.9}{ovary}).}
  \label{fig:Exemplar_frames_instrument_anatomy}
\end{figure*}

Table~\ref{tab:Timeline_visualization} presents a temporal breakdown of annotated actions across ten representative laparoscopic procedures, with each action visualized using a distinct color. The visualization highlights the significant class imbalance typical of surgical workflows, as individual videos usually contain only two to three of the defined actions. Additionally, bleeding and smoke, common side effects that often accompany primary actions, are plotted separately to illustrate their temporal overlap with core surgical activities.

\subsection{Semantic Segmentation Dataset}
\label{sec:segmentation}

The semantic segmentation component of GynSurg comprises 15 laparoscopic hysterectomy videos obtained from the Medical University of Toronto. Of these, ten videos were annotated for surgical instruments and five for anatomical structures. All annotations were performed by clinical experts following standardized guidelines to ensure high-quality, pixel-level precision.

\subsubsection{Instrument Segmentation}

The instrument segmentation dataset includes 10 annotated videos, totaling 11,352 frames at a resolution of $750 \times 480$ pixels. A total of 21 distinct instrument classes were labeled. Figure~\ref{fig:Statistics_instrument_and_anatomies} (left) shows the distribution of instrument annotations per frame. Multiple instruments often appear within the same frame, with graspers being the most frequently duplicated.

For analysis, the instruments are categorized into two functional groups:
\begin{itemize}
    \item \textbf{Primary Surgical Instruments (13 classes):} Devices directly involved in tissue manipulation, cutting, coagulation, and suturing, including grasper, suture carrier, scissor, irrigator, clip applier, needle, trocar, needle holder, corkscrew, bipolar forcep, knot pusher, sealer-divider, and hook.
    \item \textbf{Auxiliary Tools (8 classes):} Supportive devices and materials such as colpotomizer, clip, cannula, thread, trocar sleeve, morcellator, thread fragment, and glove.
\end{itemize}

Representative annotated frames illustrating both instrument and anatomy segmentation overlays are shown in Figure~\ref{fig:Exemplar_frames_instrument_anatomy}.

\subsubsection{Anatomy Segmentation}

The anatomical segmentation dataset comprises five laparoscopic hysterectomy videos, yielding 1,010 labeled frames at a resolution of $750 \times 480$ pixels. Four anatomical structures were annotated: \textit{uterus}, \textit{ovary}, \textit{fallopian tube}, and a general \textit{organ} category. For quantitative evaluation, only the uterus, ovary, and tube classes are retained due to their frequent occurrence and clinical significance. Figure~\ref{fig:Statistics_instrument_and_anatomies} (right) displays the per-frame distribution of anatomical labels, with sample annotated images presented in the rightmost panels of Figure~\ref{fig:Exemplar_frames_instrument_anatomy}.

\subsection{Unannotated Videos}
\label{sec:unannotated}

In addition to the annotated datasets, GynSurg includes a collection of 75 unannotated high-definition ($1920 \times 1080$) laparoscopic hysterectomy videos (LHE), also acquired at the University of Toronto. These recordings are provided to support research in semi-supervised and self-supervised learning for surgical video analysis.

\paragraph{Usage Notes}
The datasets are licensed under \href{https://creativecommons.org/licenses/by-nc-nd/4.0/}{CC BY-NC-ND}.
We provide all code for mask creation as well as the training IDs for four-fold validation and usage instructions in the GitHub repository of the paper \url{https://github.com/Sahar-Nasiri/GynSurg}.

\section{Methods}
\label{sec:methods}

\subsection{Action Recognition}

Developing robust methods for temporally recognizing surgical actions in gynecologic laparoscopic videos presents several challenges. First, as shown in Table~\ref{tab:dataset_composition}, the dataset is highly imbalanced. Actions such as \textit{needle passing} dominate both in case count (48 cases) and total duration (7,036 seconds), whereas actions like \textit{suction/irrigation}, despite occurring with similar frequency, span only brief intervals. Second, intraoperative visual degradations--including smoke, lens fogging, blood-induced blur, occlusions, and abrupt camera movements--can obscure critical scene elements. Finally, variability in patient anatomy, surgical workflows, and surgeon proficiency introduces diverse visual and temporal patterns, complicating model generalization across procedures.

For training and evaluation, videos are divided into three-second clips for both action and side-effect recognition tasks. Action recognition is formulated as a five-class classification problem: \textit{coagulation}, \textit{needle passing}, \textit{suction/irrigation}, \textit{transection}, and \textit{rest} (no active instrumentation). Side effect recognition is framed as two independent binary classification tasks: \textit{bleeding} vs. \textit{non-bleeding} and \textit{smoke} vs. \textit{non-smoke}. To address class imbalance, balanced sampling is used to ensure uniform class distribution within each mini-batch. 

\paragraph{Architectures.}
We evaluate four architectures designed to model the spatial and temporal dynamics of surgical videos. The first model integrates a VGG16 backbone (17.89M parameters) with a lightweight Transformer head (4.30M) for long-range feature aggregation~\cite{nasirihaghighi2024event}. The second adopts a classical CNN–RNN design, coupling VGG16 with an LSTM module (1.33M) to capture temporal dependencies. The third model replaces VGG16 with a ResNet50 backbone (24.16M), paired with the same LSTM configuration~\cite{nasirihaghighi2023action}. Lastly, we evaluate ResNet3D-18~\cite{ResNet3D} (33.24M), a fully spatiotemporal model employing 3D convolutions for joint spatial-temporal feature learning.

\paragraph{Training. }All models are trained for 40 epochs with a batch size of 16, using stochastic gradient descent (SGD) with momentum 0.9 and an initial learning rate of 0.005, which is progressively reduced during training using a polynomial decay schedule:
\[
    lr = lr_{init} \times \left(1 - \frac{iter}{total\_iter}\right)^{0.9}
\]
For networks with pre-trained backbones, the backbone’s learning rate is set to one-tenth of the main learning rate. Input frames are resized to $256\times256$ pixels, and data augmentation includes random rotations (±15°), color jittering (brightness 0.3, contrast 0.3, saturation 0.5), and Gaussian blurring (kernel size 5, $\sigma \in [0.1, 2.0]$). Models are optimized using cross-entropy loss over 30 epochs and evaluated via four-fold cross-validation; results are reported as the average across folds.

\subsection{Semantic Segmentation}

Semantic segmentation in laparoscopic videos is particularly challenging due to the variability and complexity of the visual environment. Anatomical structures such as the uterus, fallopian tubes, and ovaries vary in color, shape, size, and texture due to tissue deformation. Furthermore, rapid motion blur, specular reflections, partial occlusions, and similar appearances among different surgical instruments complicate the task.

\paragraph{Architectures. }
We benchmarke several state-of-the-art segmentation models, including DeepLabV3~\cite{DeepLabV3}, Deep Pyramid~\cite{ghamsarian2022deeppyramid}, UNet~\cite{U-Net}, UNet++~\cite{UNet++}, UPerNet~\cite{UPerNet}, CE-Net~\cite{CE-Net}, CPFNet~\cite{CPFNet}, Recal-Net~\cite{ReCal-Net}, Adapt-Net~\cite{Adapt-Net}, and the Segment Anything Model (SAM)~\cite{segment-anything}. Except for SAM, the networks is instantiated with either a VGG16 or ResNet34 backbone, pretrained on ImageNet.

\paragraph{Segment Anything Model Setting. }
We explore two fine-tuning strategies for the Segment Anything Model (SAM). In the first approach, we fine-tune only the mask decoder while keeping the vision and prompt encoders frozen, amounting to 4.05 M trainable parameters. In the second, we apply low-rank adaptation (LoRA)~\cite{hu2022lora} to the linear and convolutional layers of the vision encoder, allowing efficient fine-tuning with only 6.64 M trainable parameters. For prompting, we investigate two separate strategies. The first, which is referred to as \textit{grid prompting}, overlays a uniform grid of points on each image without prior knowledge of object locations, providing dense, weak supervision for mask generation. The second, \textit{center-point prompting}, computes the centroid of each ground-truth mask and uses it as a single, object-specific prompt, offering stronger supervision based on known object locations. 

\paragraph{Training. }Training is conducted on $256\times256$ pixel images using SGD (momentum 0.9) with an initial learning rate of 0.005. Data augmentations includes color jittering (brightness=0.5, contrast=0.5, saturation=0.5, hue=0.25) and Gaussian blurring ($\sigma=0.5$). Four-fold cross-validation is employed, with performance averaged across folds. Models are trained for 80 epochs with a batch size of 4, minimizing pixel-wise cross-entropy loss, defined as:
\begin{equation}
\begin{split}
    \mathcal{L} = (\lambda)\times CE(\mathcal{X}_{true}(i,j),\mathcal{X}_{pred}(i,j)) \\
    -(1-\lambda)\times \left( \log \frac{2\sum \mathcal{X}_{true}\odot \mathcal{X}_{pred}+\sigma}{\sum \mathcal{X}_{true} + \sum \mathcal{X}_{pred}+ \sigma} \right)
\end{split}
\label{eq:loss}
\end{equation}

\noindent Where $\mathcal{X}_{true}$ represents the ground truth mask, and $\mathcal{X}_{pred}$ denotes the predicted mask, constrained such that $0 \leq \mathcal{X}_{pred}(i,j) \leq 1$.
The weighting parameter $\lambda \in [0,1]$ is set to 0.8 in our experiments.
The symbol $\odot$ indicates the Hadamard product (element-wise multiplication), and $\sigma$ is the Laplacian smoothing constant set to 1 to prevent numerical instability and mitigate overfitting.

\begin{table}[t!]
\centering
\caption{Action recognition performance of different neural networks architectures}
\label{tab:action_recognition}
\resizebox{\columnwidth}{!}{%
\large
\begin{tabular}{lm{1.8cm}*{5}{>{\centering\arraybackslash}m{1.8cm}}}
\specialrule{.12em}{.05em}{.05em}
 \textbf{Network} &\textbf{Coagulation} & \textbf{NeedlePass.} & \textbf{Suct./Irrig.} & \textbf{Transection} & \textbf{Rest} & \textbf{Average} \\ 
 \specialrule{.12em}{.05em}{.05em}
VGG-LSTM& 51.72 \std{$\pm$ 10.30} & 93.10 \std{$\pm$ 7.46} & 61.38 \std{$\pm$ 26.91} & 68.36 \std{$\pm$ 3.49} & 48.76 \std{$\pm$ 12.25} & 68.40 \std{$\pm$ 1.76} \\
ResNet-LSTM&56.14 \std{$\pm$ 28.14} & 92.74 \std{$\pm$ 9.36} & 74.81 \std{$\pm$ 29.01} & 84.28 \std{$\pm$ 6.24} & 56.82 \std{$\pm$ 11.21} & 76.99 \std{$\pm$ 6.69} \\
VGG-Transformer &49.13 \std{$\pm$ 10.19} & 94.68 \std{$\pm$ 5.37} & 62.79 \std{$\pm$ 24.77} & 75.64 \std{$\pm$ 6.08} & 24.33 \std{$\pm$ 22.53} & 69.29 \std{$\pm$ 1.01} \\
ResNet3D&45.11 \std{$\pm$ 20.26} & 99.54 \std{$\pm$ 0.61} & 91.51 \std{$\pm$ 13.42} & 12.27 \std{$\pm$ 12.09} & 57.15 \std{$\pm$ 17.30} & 58.46 \std{$\pm$ 5.67} \\
\specialrule{.12em}{.05em}{.05em}
\end{tabular}
}
\end{table}

\begin{table}[t!]
\centering
\caption{Side-effects recognition performance of different neural networks architectures}
\label{tab:side_effect_recognition}
\resizebox{\columnwidth}{!}{%
\begin{tabular}{lm{2cm}*{3}{>{\centering\arraybackslash}m{2cm}}}
\specialrule{.12em}{.05em}{.05em}
 & \multicolumn{2}{c}{\textbf{Bleeding}} & \multicolumn{2}{c}{\textbf{Non-bleeding}} \\ \cmidrule(lr){2-3}\cmidrule(lr){4-5}
 Network & Accuracy (\%) & F1-Score (\%) & Accuracy (\%) & F1-Score (\%) \\ \specialrule{.12em}{.05em}{.05em}
VGG-LSTM& 84.78 \std{$\pm$ 8.41} & 86.09 \std{$\pm$ 3.85} & 88.89 \std{$\pm$ 5.00} & 85.16 \std{$\pm$ 8.89} \\
ResNet-LSTM&88.26 \std{$\pm$ 2.01} & 92.51 \std{$\pm$ 0.92} & 97.04 \std{$\pm$ 2.12} & 92.38 \std{$\pm$ 4.23} \\
VGG-Transformer & 85.17 \std{$\pm$ 7.63} & 88.11 \std{$\pm$ 3.53} & 91.81 \std{$\pm$ 3.69} & 87.06 \std{$\pm$ 9.73}\\
ResNet3D&90.36 \std{$\pm$ 2.57} & 84.89 \std{$\pm$ 6.08} & 79.68 \std{$\pm$ 10.36} & 83.05 \std{$\pm$ 7.04} \\
\end{tabular}
}

\resizebox{\columnwidth}{!}{%
\begin{tabular}{lm{2cm}*{3}{>{\centering\arraybackslash}m{2cm}}}
\specialrule{.12em}{.05em}{.05em}
 & \multicolumn{2}{c}{\textbf{Smoke}} & \multicolumn{2}{c}{\textbf{Non-smoke}} \\ \cmidrule(lr){2-3}\cmidrule(lr){4-5}
 Network & Accuracy (\%) & F1-Score (\%) & Accuracy (\%) & F1-Score (\%) \\ \specialrule{.12em}{.05em}{.05em}
VGG-LSTM& 72.86 \std{$\pm$ 9.42} & 77.92 \std{$\pm$ 3.72} & 87.77 \std{$\pm$ 8.61} & 81.30 \std{$\pm$ 7.33} \\
ResNet-LSTM&80.69 \std{$\pm$ 5.33} & 85.58 \std{$\pm$ 3.44} & 91.36 \std{$\pm$ 7.05} & 86.67 \std{$\pm$ 6.78} \\
VGG-Transformer &73.81 \std{$\pm$ 7.81} & 80.18 \std{$\pm$ 2.90} & 91.41 \std{$\pm$ 6.33} & 83.79 \std{$\pm$ 6.30}\\
ResNet3D&83.54 \std{$\pm$ 4.77} & 86.18 \std{$\pm$ 2.11} & 88.05 \std{$\pm$ 9.63} & 86.02 \std{$\pm$ 6.70}\\

\specialrule{.12em}{.05em}{.05em}
\end{tabular}
}
\end{table}

\begin{table}[t!]
\centering
\caption{Dice coefficient (\%) for auxiliary tool segmentation across different neural network architectures}
\label{tab:aux_tool_segmentation}
\resizebox{\columnwidth}{!}{%
\begin{tabular}{lm{1.3cm}*{5}{>{\centering\arraybackslash}m{1.8cm}}}
\specialrule{.12em}{.05em}{.05em}
\textbf{Backbone} &  \textbf{Network} & \textbf{Morcellator} & \textbf{Thread} & \textbf{Trocar-sleeve} & \textbf{Cannula} & \textbf{Average} \\ \specialrule{.12em}{.05em}{.05em}
\multirow{6}{*}{VGG16}
&DeepPyramid& 89.28 \std{$\pm$ 1.65} & 71.62 \std{$\pm$ 1.74} & 12.29 \std{$\pm$ 9.20} & 87.46 \std{$\pm$ 2.48} & 65.16 \std{$\pm$ 2.71} \\
&DeepLabV3&86.12 \std{$\pm$ 1.43} & 66.85 \std{$\pm$ 3.56} & 19.20 \std{$\pm$ 8.77} & 89.85 \std{$\pm$ 3.35} & 65.50 \std{$\pm$ 3.17}\\
&UNet&87.95 \std{$\pm$ 2.85} & 72.04 \std{$\pm$ 1.28} & 11.02 \std{$\pm$ 4.11} & 63.45 \std{$\pm$ 12.11} & 58.62 \std{$\pm$ 4.50}\\
&CE-Net &83.50 \std{$\pm$ 3.97} & 68.21 \std{$\pm$ 2.18} & 14.80 \std{$\pm$ 5.29} & 85.43 \std{$\pm$ 4.17} & 60.49 \std{$\pm$ 2.33}\\
&CPFNet &88.61 \std{$\pm$ 2.37} & 68.55 \std{$\pm$ 2.31} & 10.95 \std{$\pm$ 5.37} & 72.80 \std{$\pm$ 10.54} & 60.23 \std{$\pm$ 2.30}\\
&RecalNet & 89.22 \std {${\pm}$ 2.32} & 72.47 \std {${\pm}$ 1.86} & 12.47 \std {${\pm}$ 9.38} & 27.46 \std {${\pm}$ 6.13} & 50.41 \std {${\pm 2.78}$} \\

\midrule\multirow{2}{*}{ResNet34}
&DeepPyramid&84.46 \std{$\pm$ 2.26} & 68.92 \std{$\pm$ 0.76} & 15.29 \std{$\pm$ 2.04} & 80.11 \std{$\pm$ 5.15} & 59.70 \std{$\pm$ 1.05}\\
&CPFNet &73.27 \std{$\pm$ 5.28} & 68.88 \std{$\pm$ 2.24} & 12.56 \std{$\pm$ 1.31} & 41.95 \std{$\pm$ 13.12} & 46.66 \std{$\pm$ 1.73} \\
\specialrule{.12em}{.05em}{.05em}
\end{tabular}%
}
\end{table}

\paragraph{Instrument Segmentation Setting.} Originally, 13 distinct instrument classes were annotated. To avoid overfitting, three underrepresented classes (trocar, clip applier, corkscrew) are excluded. The four instruments related to suturing (suture-carrier, knot-pusher, needle-holder, needle) are merged into a single \textit{suturing instrument} class. This results in seven final classes: grasper, scissors, irrigator, bipolar forceps, sealer-divider, hook, and suturing instrument.

\begin{figure}[t!]
    \centering    
    \includegraphics[width=0.45\textwidth]{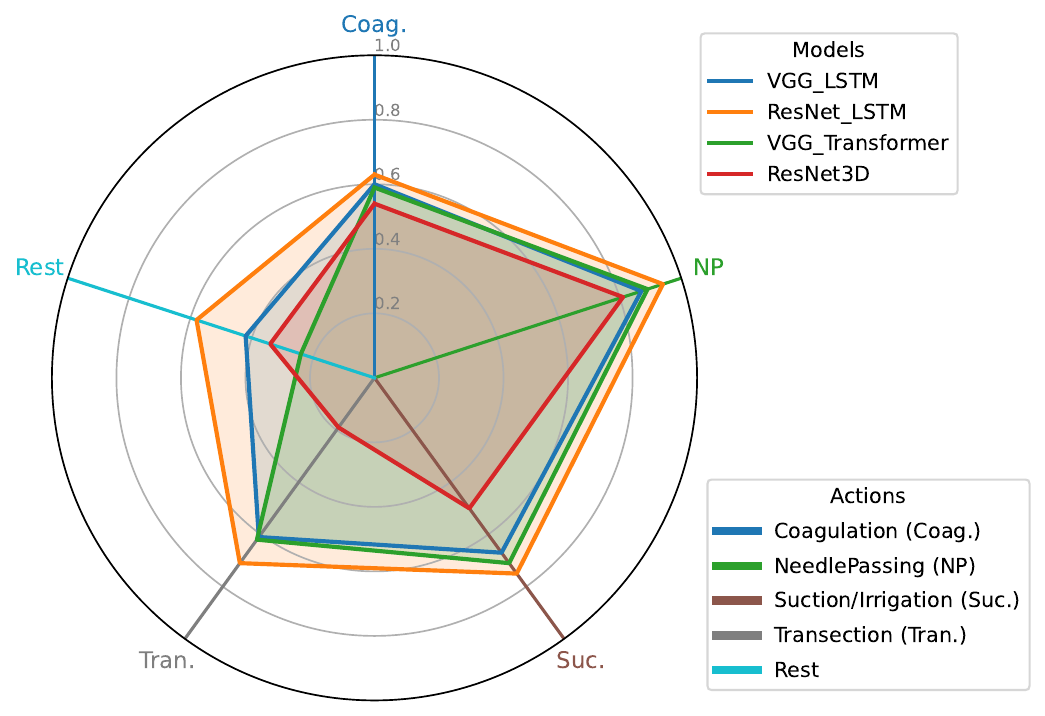}    
    \caption{F1‐scores for each action across different models}
    \Description{F1‐scores for each action across different models}
    \label{fig:F1score action recognition}
\end{figure}

\paragraph{Auxiliary Tool Segmentation Setting.} Eight auxiliary tools were initially annotated. Due to limited cases, clip, colpotomizer, and glove are excluded. Cannula and in-cannula are merged into a single \textit{cannula} category; thread and thread-fragment are merged into \textit{thread}. The final set includes four auxiliary tool classes: thread, trocar-sleeve, cannula, and morcellator.

\paragraph{Anatomy Segmentation Setting.} Four anatomical structures were initially labeled: uterus, fallopian tube, ovary, and a generic \textit{organ} class. Due to ambiguity and low frequency, the organ category is excluded from evaluation. The final segmentation targets are the uterus, fallopian tube, and ovary, providing clinically relevant, high-quality segmentations for context-aware assistance.

\section{Experimental Results}
\label{sec:experimental results}
In this section, we evaluate our methods on a gynecological laparoscopic surgery dataset, with comprehensive results for action recognition and side effect detection, surgical instruments, auxiliary tools, and anatomical structures segmentation detailed below.

\subsection{Action Recognition}
In this section, we evaluate four architectures, VGG-LSTM, ResNet-LSTM, VGG-Transformer, and ResNet3D, on our surgical action and side effect dataset. Table~\ref{tab:action_recognition} reports each model’s accuracy across all action classes: Coagulation, Transection, Needle Passing, Suction/Irrigation, and Rest.

\begin{table*}[th!]
\centering
\caption{Dice coefficient (\%) for instrument segmentation across different neural network architectures}
\label{tab:instrument_segmentation}
\resizebox{\textwidth}{!}{%
\begin{tabular}{lm{1.6cm}*{8}{>{\centering\arraybackslash}m{2cm}}}
\specialrule{.12em}{.05em}{.05em}
\textbf{Backbone} &  \textbf{Network} & \textbf{Grasper} & \textbf{Scissors} & \textbf{Irrigator} & \textbf{Bipolar-forc.} & \textbf{Sealer-divider} & \textbf{Hook} & \textbf{Suturing} & \textbf{Average} \\ \specialrule{.12em}{.05em}{.05em}
\multirow{7}{*}{VGG16}
&DeepPyramid& 76.04 \std{$\pm$ 3.02} & 51.62 \std{$\pm$ 7.92} & 69.27 \std{$\pm$ 4.72} & 84.30 \std{$\pm$ 3.14} & 84.70 \std{$\pm$ 3.44} & 71.35 \std{$\pm$ 7.78} & 77.04 \std{$\pm$ 2.58} & 73.47 \std{$\pm$ 1.94} \\
&DeepLabV3&77.42 \std{$\pm$ 2.41} & 57.62 \std{$\pm$ 3.72} & 75.57 \std{$\pm$ 4.63} & 83.35 \std{$\pm$ 2.10} & 85.73 \std{$\pm$ 1.72} & 72.28 \std{$\pm$ 9.90} & 77.51 \std{$\pm$ 3.30} & 75.64 \std{$\pm$ 2.30} \\
&UNet++ &73.51 \std{$\pm$ 3.60} & 41.14 \std{$\pm$ 9.76} & 66.37 \std{$\pm$ 6.75} & 80.48 \std{$\pm$ 1.63} & 83.67 \std{$\pm$ 1.76} & 62.79 \std{$\pm$ 6.76} & 68.77 \std{$\pm$ 1.41} & 68.10 \std{$\pm$ 1.81} \\
&UNet&73.84 \std{$\pm$ 2.42} & 40.48 \std{$\pm$ 9.74} & 68.14 \std{$\pm$ 6.43} & 80.20 \std{$\pm$ 1.99} & 83.75 \std{$\pm$ 2.04} & 62.37 \std{$\pm$ 10.05} & 69.94 \std{$\pm$ 2.16} & 68.39 \std{$\pm$ 3.49} \\
&UPerNet &76.92 \std{$\pm$ 2.96} & 54.61 \std{$\pm$ 6.02} & 74.72 \std{$\pm$ 5.69} & 83.46 \std{$\pm$ 0.50} & 85.50 \std{$\pm$ 1.94} & 65.85 \std{$\pm$ 6.65} & 76.00 \std{$\pm$ 3.16} & 73.87 \std{$\pm$ 2.07} \\
&CE-Net &73.87 \std{$\pm$ 3.31} & 53.00 \std{$\pm$ 9.52} & 69.34 \std{$\pm$ 6.61} & 82.41 \std{$\pm$ 2.44} & 79.86 \std{$\pm$ 2.03} & 70.89 \std{$\pm$ 9.64} & 73.03 \std{$\pm$ 4.28} & 71.77 \std{$\pm$ 3.95} \\
&CPFNet &75.16 \std{$\pm$ 3.40} & 53.12 \std{$\pm$ 8.63} & 72.00 \std{$\pm$ 6.01} & 82.14 \std{$\pm$ 1.55} & 82.82 \std{$\pm$ 2.79} & 65.96 \std{$\pm$ 10.96} & 75.26 \std{$\pm$ 4.00} & 72.35 \std{$\pm$ 3.34} \\
&RecalNet & 74.50 \std{$\pm$ 2.30} & 51.15 \std {${\pm}$ 11.28} & 66.74 \std {${\pm}$ 5.08} & 80.48 \std {${\pm}$ 3.91} & 82.84 \std {${\pm}$ 3.49} & 65.94 \std {${\pm}$ 10.69} & 73.56 \std {${\pm}$ 3.98} & 70.74 \std {${\pm 3.09}$} \\
&AdaptNet & 73.99\std {${\pm}$ 3.64} & 47.89\std {${\pm}$ 6.34} & 65.29\std {${\pm}$ 4.58} & 82.62\std {${\pm}$ 3.48} & 83.42\std {${\pm}$ 2.98} & 55.76\std {${\pm}$ 19.02} & 71.88\std {${\pm}$ 4.96} & 68.69\std {${\pm 4.35}$} \\

\midrule\multirow{5}{*}{ResNet34}
&DeepPyramid&74.39 \std{$\pm$ 2.49} & 54.30 \std{$\pm$ 8.16} & 70.03 \std{$\pm$ 2.90} & 82.60 \std{$\pm$ 3.44} & 82.33 \std{$\pm$ 4.08} & 73.16 \std{$\pm$ 5.85} & 73.05 \std{$\pm$ 2.99} & 72.84 \std{$\pm$ 1.68} \\
&UNet++ &71.61 \std{$\pm$ 3.17} & 46.24 \std{$\pm$ 4.66} & 61.90 \std{$\pm$ 5.48} & 81.22 \std{$\pm$ 2.96} & 84.88 \std{$\pm$ 1.59} & 52.46 \std{$\pm$ 13.55} & 68.73 \std{$\pm$ 1.81} & 66.72 \std{$\pm$ 1.89} \\
&UNet&72.33 \std{$\pm$ 1.97} & 38.98 \std{$\pm$ 0.10} & 61.22 \std{$\pm$ 0.93} & 78.19 \std{$\pm$ 3.38} & 77.02 \std{$\pm$ 1.23} & 13.53 \std{$\pm$ 10.49} & 62.32 \std{$\pm$ 7.60} & 57.66 \std{$\pm$ 1.15} \\
&CE-Net &68.75 \std{$\pm$ 0.73} & 33.15 \std{$\pm$ 0.97} & 60.01 \std{$\pm$ 0.08} & 74.94 \std{$\pm$ 3.76} & 70.31 \std{$\pm$ 9.32} & 21.27 \std{$\pm$ 20.98} & 58.64 \std{$\pm$ 0.14} & 55.30 \std{$\pm$ 1.16} \\
&CPFNet &71.30 \std{$\pm$ 0.24} & 45.53 \std{$\pm$ 6.91} & 68.68 \std{$\pm$ 2.84} & 64.46 \std{$\pm$ 9.01} & 84.56 \std{$\pm$ 1.66} & 69.42 \std{$\pm$ 6.80} & 69.27 \std{$\pm$ 3.19} & 67.60 \std{$\pm$ 2.37} \\
\specialrule{.12em}{.05em}{.05em}
\end{tabular}
}
\end{table*}

As shown, ResNet-LSTM achieves the highest average performance. All methods exceed 90\% accuracy on Needle Passing, reflecting both its high frequency in the dataset and the use of distinctive suturing instruments. In contrast, Coagulation and Transection are frequently misclassified, as both use similar instruments that generate smoke, and the relative scarcity of Transection cases further lowers its detection accuracy. Finally, the Rest category, comprising brief intervals immediately before or after any action, often depicts the same instruments seen in other classes, making it inherently difficult to distinguish. Figure~\ref{fig:F1score action recognition} compares F1‐scores for each model across all classes. 

Table~\ref{tab:side_effect_recognition} reports binary classification results for side-effect detection (bleeding vs. non-bleeding, smoke vs. non-smoke) across the four evaluated architectures. ResNet–LSTM achieves the highest performance in both tasks, with an accuracy of 92.65\% for bleeding detection and 86.03\% for smoke classification, closely followed by ResNet3D with 85.80\% accuracy in the latter.

\subsection{Semantic Segmentation}
For semantic segmentation, we evaluate multiple architectures on datasets of surgical instruments, auxiliary tools, and anatomical structures, using the Dice coefficient as our primary performance metric.

Tables~\ref{tab:aux_tool_segmentation}--\ref{tab:sam_seg} summarize segmentation performance across auxiliary tools, instruments, anatomical structures, and SAM variants. For auxiliary tools, DeepLabV3 with a VGG16 backbone achieve the highest mean Dice score (65.50\%), with DeepPyramid excelling on morcellator (89.28\%) and DeepLabV3 leading on cannula (89.85\%). Trocar-sleeve segmentation remains challenging across models, likely due to limited annotations and small object size.

In instrument segmentation (Table~\ref{tab:instrument_segmentation}), DeepLabV3 again leads with an average Dice score of 75.64\%, with bipolar-forceps and sealer-divider classes consistently exceeding 80\% due to their distinct visual features.
Anatomy segmentation results (Table~\ref{tab:anatomy segmentation}) are generally lower, reflecting the deformable and variable nature of tissues. Ovary segmentation achieves the highest Dice scores, benefiting from more consistent morphology compared to uterus and tube.

Finally, Table~\ref{tab:sam_seg} compares three SAM variants: vanilla SAM, SAM-LoRA, and SAM with center-point prompting (SAM-PP), across five classes (uterus, ovary, forceps, thread, morcellator). Both the vanilla SAM and the LoRA-adapted SAM are fine-tuned using grid-based prompts; SAM-LoRA consistently outperforms the unmodified model by employing low-rank parameter updates that better capture the subtle textures and boundaries unique to laparoscopic imagery. In contrast, applying a single center-point prompt to the original SAM (SAM-PP) yields the highest segmentation accuracy, achieving Dice scores up to 90\% for morcellator, by providing precise object-location cues at inference time.

\begin{table}[t!]
\centering
\caption{Dice coefficient (\%) for anatomy segmentation across different neural network architectures}
\label{tab:anatomy segmentation}
\resizebox{\columnwidth}{!}{%

\begin{tabular}{lm{1.6cm}*{4}{>{\centering\arraybackslash}m{2cm}}}
\specialrule{.12em}{.05em}{.05em}
\textbf{Backbone} &  \textbf{Network} & \textbf{Uterus} & \textbf{Tube} & \textbf{Ovary} & \textbf{Average} \\ \specialrule{.12em}{.05em}{.05em}
\multirow{7}{*}{VGG16}
&DeepPyramid& 63.29 \std{$\pm$ 16.06} & 16.83 \std{$\pm$ 8.90} & 72.52 \std{$\pm$ 2.66} & 50.88 \std{$\pm$ 7.04} \\
&DeepLabV3&64.35 \std{$\pm$ 10.77} & 18.94 \std{$\pm$ 9.87} & 71.58 \std{$\pm$ 4.11} & 51.62 \std{$\pm$ 6.12} \\
&UNet++ &58.79 \std{$\pm$ 13.53} & 18.43 \std{$\pm$ 12.65} & 71.58 \std{$\pm$ 3.98} & 49.60 \std{$\pm$ 6.81} \\
&UNet&57.17 \std{$\pm$ 14.42} & 20.34 \std{$\pm$ 12.38} & 70.98 \std{$\pm$ 4.20} & 49.49 \std{$\pm$ 7.08}\\
&UPerNet &65.06 \std{$\pm$ 14.81} & 20.92 \std{$\pm$ 12.70} & 70.25 \std{$\pm$ 5.29} & 52.08 \std{$\pm$ 8.62} \\
&CE-Net &57.90 \std{$\pm$ 15.31} & 16.04 \std{$\pm$ 8.14} & 69.48 \std{$\pm$ 3.69} & 47.80 \std{$\pm$ 7.60} \\
&CPFNet &56.20 \std{$\pm$ 16.26} & 17.93 \std{$\pm$ 11.98} & 69.36 \std{$\pm$ 3.83} & 47.83 \std{$\pm$ 8.77} \\
&RecalNet & 63.02 \std {${\pm}$ 11.97} & 18.34 \std {${\pm}$ 9.95} & 71.80 \std {${\pm}$ 4.08} & 51.05 \std {${\pm 5.30}$} \\

\midrule\multirow{5}{*}{ResNet34}
&DeepPyramid&61.50 \std{$\pm$ 13.27} & 18.62 \std{$\pm$ 12.71} & 67.80 \std{$\pm$ 5.10} & 49.31 \std{$\pm$ 7.63} \\
&UNet++ &61.84 \std{$\pm$ 17.10} & 20.83 \std{$\pm$ 9.66} & 73.10 \std{$\pm$ 5.24} & 51.92 \std{$\pm$ 7.25} \\
&UNet&58.32 \std{$\pm$ 13.08} & 16.14 \std{$\pm$ 13.52} & 68.61 \std{$\pm$ 5.91} & 47.69 \std{$\pm$ 8.07} \\
&CE-Net &61.12 \std{$\pm$ 12.76} & 22.31 \std{$\pm$ 6.61} & 65.78 \std{$\pm$ 9.45} & 49.73 \std{$\pm$ 7.52} \\
&CPFNet &60.95 \std{$\pm$ 16.07} & 19.41 \std{$\pm$ 9.05} & 68.30 \std{$\pm$ 4.21} & 49.55 \std{$\pm$ 6.34} \\

\specialrule{.12em}{.05em}{.05em}
\end{tabular}
}
\end{table}
\begin{table}[t!]
\centering
\caption{Dice coefficient (\%) for instrument and anatomy segmentation using different SAM model variants.}
\label{tab:sam_seg}
\resizebox{\columnwidth}{!}{%
\begin{tabular}{lm{1.3cm}*{5}{>{\centering\arraybackslash}m{1.8cm}}}
\specialrule{.12em}{.05em}{.05em}
 \textbf{Model} & \textbf{Uterus} & \textbf{Ovary} & \textbf{Bipolar-forc.} & \textbf{Thread} & \textbf{Morcellator} \\ \specialrule{.12em}{.05em}{.05em}
SAM& 49.32 \std {${\pm 5.03}$} & 57.63 \std {${\pm 9.95}$} & 47.40 \std {${\pm 6.94}$} & 56.82 \std {${\pm 1.54}$} & 45.09 \std {${\pm 9.24}$} \\
SAM-LoRA& 55.80 \std {${\pm 3.86}$} & 69.59 \std {${\pm 11.90}$} & 64.09 \std {${\pm 4.94}$} & 64.30 \std {${\pm 1.35}$} & 63.64 \std {${\pm 5.34}$}\\
SAM-PP &85.72 \std{${\pm 1.62}$} & 82.73 \std{${\pm 3.44}$} & 81.63 \std{${\pm 3.40}$} & 73.91\std{${\pm 2.33}$} & 90.48 \std{${\pm3.11}$}\\
\specialrule{.12em}{.05em}{.05em}
\end{tabular}%
}
\end{table}

\section{Conclusion}
\label{sec:Conclusion}

In summary, we introduce GynSurg, the largest multi‐task gynecologic laparoscopy dataset, comprising 152 high‐definition videos annotated for four operative actions and two intraoperative side effects, 12362 frames with pixel‐level masks for 21 instruments and three key organs, plus 75 unannotated hysterectomy recordings. Through standardized benchmarks of four action recognition models and several segmentation architectures (including segment anything model), we establish reproducible performance baselines and highlight persistent challenges such as class imbalance, visual noise, and anatomical variability. The dataset and annotations are publicly available to support advancements in surgical training and postoperative video analysis.

\begin{acks} 
This work was partly funded by the FWF Austrian Science Fund under grant P~32010-N38. We would like to thank Sabrina Kletz for the supervision and processing of instrument annotations. We further thank Daniela Fox for annotating videos. 
\end{acks}

\bibliographystyle{acm}
\bibliography{references}

\end{document}